%% file: aors_ruiz_sell.tex
\documentclass{article}

\usepackage[nonatbib,final]{neurips_2023}

\makeatletter
\renewcommand{\@notice}{}
\makeatother

\usepackage[utf8]{inputenc} 
\usepackage[T1]{fontenc}    
\usepackage{hyperref}       
\usepackage{url}            
\usepackage{booktabs}       
\usepackage{colortbl}       
\usepackage{amsfonts}       
\usepackage{nicefrac}       
\usepackage{microtype}      
\usepackage{xcolor}         
\usepackage{indentfirst}    
\usepackage{graphicx}       
\usepackage{caption}        
\usepackage{subcaption}     
\usepackage{float}          
\usepackage{fancyvrb}       
\usepackage{fvextra}        
\RecustomVerbatimEnvironment{Verbatim}{Verbatim}{breaklines=true, breakanywhere=true}
\usepackage{algorithm}
\usepackage{algorithmic}
\usepackage{milbench}
\usepackage{textcomp}

\DefineVerbatimEnvironment{verbatim}{Verbatim}{breaklines=true}

\makeatletter
\def\ps@firstpage{%
  \let\@oddhead\@empty
  \let\@evenhead\@empty
  \def\@oddfoot{%
    \hfil 
    \small 
    Distribution Statement A. Approved for public release: distribution is unlimited.%
    \hfil} 
  \let\@evenfoot\@oddfoot
}
\makeatother

\begin{document}

\title{Fine-Tuning and Evaluating Open-Source Large Language Models for the Army Domain}

%

\author{
  MAJ Daniel C. Ruiz\footnotemark[1] \\
  US Army Futures Command\\
  The Research and Analysis Center\\
  Monterey, CA 93940\\
  \And
  John Sell\footnotemark[1] \\
  US Army Futures Command\\
  The Research and Analysis Center\\
  Fort Leavenworth, KS 66027 \\
}
\maketitle

\thispagestyle{firstpage}

\renewcommand{\thefootnote}{\fnsymbol{footnote}}
\footnotetext[1]{Equal Contribution}
\renewcommand{\thefootnote}{\arabic{footnote}}


\input{draft_content/0_buildup}

\input{draft_content/1_traclm}

\input{draft_content/2_milbench}

\input{draft_content/3_results}

\input{draft_content/4_winddown}

\newpage

\bibliography{aors_ruiz_sell}{}
\bibliographystyle{plain}

\newpage

\input{draft_content/5_appendix}

\end{document}

%% file: draft_content/0_buildup.tex
\begin{abstract}
In recent years, the widespread adoption of Large Language Models (LLMs) has sparked interest in their potential for application within the military domain. However, the current generation of LLMs demonstrate sub-optimal performance on Army use cases, due to the prevalence of domain-specific vocabulary and jargon. In order to fully leverage LLMs in-domain, many organizations have turned to fine-tuning to circumvent the prohibitive costs involved in training new LLMs from scratch. In light of this trend, we explore the viability of adapting open-source LLMs for usage in the Army domain in order to address their existing lack of domain-specificity. Our investigations have resulted in the creation of three distinct generations of TRACLM, a family of LLMs fine-tuned by The Research and Analysis Center (TRAC), Army Futures Command (AFC). Through continuous refinement of our training pipeline, each successive iteration of TRACLM displayed improved capabilities when applied to Army tasks and use cases. Furthermore, throughout our fine-tuning experiments, we recognized the need for an evaluation framework that objectively quantifies the Army domain-specific knowledge of LLMs. To address this, we developed MilBench, an extensible software framework that efficiently evaluates the Army knowledge of a given LLM using tasks derived from doctrine and assessments. We share preliminary results, models, methods, and recommendations on the creation of TRACLM and MilBench. Our work  significantly informs the development of LLM technology across the DoD and augments senior leader decisions with respect to artificial intelligence integration. 
\end{abstract}

\section{Introduction}

\subsection{Background}

In the fast-paced fields of machine learning (ML) and natural language processing (NLP), LLMs have garnered worldwide attention as a highly transformative technology with virtually limitless applications across multiple domains. Popularized by high-impact commercial LLMs like OpenAI's ChatGPT \cite{openai2023chatgpt}, Google's Gemini \cite{geminiteam2023gemini}, and Anthropic's Claude \cite{anthropic2024claude3blogpost}, LLMs have become household terms due to their uncanny ability to emulate human-like understanding and advanced problem-solving abilities. Concurrent with the rise of LLMs from leading technology companies, open-source LLMs such as BLOOM \cite{workshop2023bloom}, LLaMA \cite{touvron2023llama}, and Mistral \cite{jiang2023mistral} have made advanced artificial intelligence freely available to researchers for application to highly specialized domains, including the U.S. Army. The advent of these powerful open-source baseline models is particularly significant, due to the well-known prohibitive cost of training an LLM from scratch \cite{Ling2023BeyondOA}. Given the existence of permissively licensed open-source LLMs, researchers and small organizations with limited resources have turned to fine-tuning in order to adapt LLMs to domains and tasks of interest. Albeit a generic term with many forms of nuanced implementation\cite{cheng2024adapting}, fine-tuning often involves training a baseline LLM on additional text deemed to adequately represent a particular domain. After successful fine-tuning, the resulting model will exhibit increased knowledge about the target domain, and potentially improved performance on domain-specific tasks. This result is especially attractive for analytical organizations, since workflows that were previously unattainable due to a lack of personnel, expertise, time, or clean data can be automated and scaled as needed. Furthermore, fine-tuning LLMs for improved performance in a target domain should be of particular interest to the U.S. Army, due to the prevalence of Army-specific vocabulary, acronyms, and jargon found within our profession. Moreover, fine-tuned LLMs that are fully hosted and controlled by the U.S. Army, as opposed to industry partners, naturally reduce operating costs as well as the risk of leaking classified operational material into commercial environments. 

\subsection{Problem Definition}

During TRAC's support for Project Convergence (PC) Capstone Four (C4) in October 2022, we first observed that an Army-specific fine-tuned LLM could benefit Army analytics. PC is a large-scale Army experiment that seeks to inform the acquisitions process by testing the compatibility and interoperability of various in-development technologies across key use cases. In recent years, TRAC provided support for PC C4 in the form of in-stride analysis, where daily snapshots of experiment output are collected and refined in near-real-time to inform long-tail analytical efforts and senior leader decisions. In support of the in-stride analysis mission, TRAC successfully leveraged T0\cite{sanh2022multitask}, a 3B-parameter open-source LLM, to provide sentiment analysis, topic modeling, and summarization of unstructured documents on classified networks. While T0 performed admirably in many cases, it became increasingly clear to TRAC analysts that model output faltered when it was asked to process text full of Army jargon and technical descriptions of Army equipment. Furthermore, swapping T0 for a commercial LLM solution would have been infeasible, both due to the classified operating environment of PC and a lack of options in the pre-ChatGPT era. Even if today's near-SOTA LLMs existed at the time, they would still suffer from a lack of domain specificity, due to being trained largely on generic internet-sourced text\cite{liu2024datasets}. In light of this problematic capability gap, we scoped our research questions as follows:
\begin{enumerate}
    \item Can fine-tuning effectively inject domain-specific Army knowledge into open-source LLMs?
    \item How can we validate that a fine-tuned LLM has successfully acquired target knowledge?
\end{enumerate}

\subsection{Contribution}

In response to our problem statement and research questions, this paper documents techniques, results, and lessons learned in both the development and evaluation of TRACLM and MilBench. Contributions include:

\begin{itemize}
\item A detailed road-map for fine-tuning three TRACLM versions, highlighting techniques that improved model performance with each iteration.
\item A technical description of MilBench, an Army-built modular LLM evaluation framework with high potential for broad application across the DoD.
\item Evaluation results of TRACLM and other open-source models using MilBench tasks and general-purpose benchmarks.
\item Descriptions of TRACLM \& MilBench technical limitations, safety and security considerations, and recommendations for future work.
\end{itemize}

\section{Related Work}

\subsection{Army Research}

To our knowledge, the TRACLM project is the first and only attempt to fine-tune open-source LLMs for the Army domain at the scale and scope discussed in this paper. While small-scale academic experiments from the Army Research Lab (ARL) and Artificial Intelligence Integration Center (AI2C) are known to have surveyed the LLM training design space, these efforts were largely exploratory and remain unpublished. However, two years prior to the popularization of LLMs through ChatGPT, an impactful collaboration between the Massachusetts Institute of Technology Lincoln Laboratory (MIT-LL) and U.S. Army Project Manager Mission Command (PMMC) resulted in the creation of MilGLUE \cite{milglue} and associated Bidirectional Encoder Representations from Transformers (BERT) \cite{bert} fine-tuned language models. Prior to the popularization of LLMs, BERT-like models\footnote{Precursors to LLMs that, despite also leveraging the transformer neural network architecture\cite{vaswaniAttentionAllYou2017}, were designed to encode text into high-dimensional representations rather than generate new text.} served as an important stepping stone in creating modern statistical approximations of natural language. Similarly, the MilGLUE project served as direct inspiration and a useful starting point for vital components of our research (see Section \ref{sec:MilBench}: MilBench).

\subsection{Academia \& Industry Research}

Despite the evident lack of Army fine-tuning examples before the TRACLM project, there has been no shortage of domain adaptation experiments in academia and industry. Domain-specific knowledge has been encoded into baseline LLMs successfully in numerous sectors, including the medical domain\cite{han2023medalpaca}, financial domain\cite{wang2023fingpt}, and other highly specialized domains such as silicon chip design\cite{liu2024chipnemo}. The success of such endeavors strongly suggests that open-source LLMs can also be calibrated for Army, military, and defense-related utilization through effective fine-tuning. Additionally, in our observation, LLM fine-tuning research generally supports either one or both of two interrelated goals: improving the performance of smaller open-source LLMs through innovative training techniques, or democratizing access to LLMs by making fine-tuning more efficient. These goals are interrelated because boosting smaller LLM performance and fine-tuning efficiency reduces hardware requirements for researchers, thus tempering a reliance on commercial, closed-source near-SOTA LLMs. 

\subsubsection{Improving Small LLM Performance}

Researchers have discovered a number of techniques in recent years that help close the gap between LLMs in the 3-7B parameter range and much larger, near-SOTA LLMs with hundreds of billions or even trillions of parameters\cite{BaGuaLu}. Many of these techniques informed our training procedures and experiments for TRACLM. These include optimal learning rates based on published LLM scaling laws\cite{hoffmann2022training}, leveraging synthetic training data derived from near-SOTA LLMs\cite{alpaca}, and teaching LLMs to follow instructions by converting raw domain text into questions and answers\cite{cheng2024adapting}. Furthermore, recent open-source research has popularized model alignment techniques like direct preference optimization (DPO)\cite{rafailov2023direct} and reinforced token optimization (RTO)\cite{zhong2024dpo}. These techniques are notable because they are viable alternatives to reinforcement learning through human feedback (RLHF)\cite{bai2022training}, thus reducing the need for active human participation in the training loop.

\subsubsection{Democratizing LLM Fine-tuning}

In service of increasing fine-tuning efficiency, particularly in terms of memory requirements, parameter efficient fine-tuning (PEFT) methods have become the norm\cite{han2024parameterefficient}. Among these methods, Dettmers et. al's Quantized Low Rank Adapters (QLoRA)\cite{dettmers2023qlora} deserves special mention for its introduction of three novel quantization techniques. When applied in tandem, these techniques enable researchers to fine-tune 70B-parameter LLMs (approximately 280GB models in 32-bit precision) on consumer-grade graphics processing units (GPUs) with minimal performance degradation. While TRAC's local hardware allowed us to fully fine-tune LLMs in the 3-7B parameter range without the need for quantization, projects like QLoRA and, more recently, model merging\cite{goddard2024arcees} suggest a near future where Army fine-tunes in the 70B+ parameter range are well within the realm of possibility. If this trend continues, we may reach an end-state where fine-tuning hardware requirements become virtually negligible, allowing any DoD organization to create LLMs for bespoke purposes on demand without relying on commercial solutions.

\subsection{Statement of Assumptions}

Given our own observations and the related work outlined above, our approach for the TRACLM + MilBench project relied upon the following assumptions:
\begin{enumerate}
\item The unclassified subset of Army doctrine provides sufficient domain-specific knowledge to noticeably improve the performance of a pre-trained LLM through fine-tuning.
\item Due to linguistic capabilities LLMs acquire through pre-training, minimal pre-processing of Army doctrine would be required to fine-tune a given LLM.
\item Open-source LLMs, as opposed to commercial alternatives, are already performant enough to warrant fine-tuning on domain-specific data.
\end{enumerate}

As we demonstrate below, these assumptions are both validated and challenged throughout the course of our research.

%% file: draft_content/1_traclm.tex
\section{TRACLM}

The TRACLM project is a first-of-its-kind effort to create a fine-tuned LLM intended for broad usage across the Army. Over the course of several months, TRAC has run extensive experiments on local hardware in partnership with the Naval Postgraduate School (NPS), resulting in three distinct generations of TRACLM models. Below, we detail the acquisition and preparation of training data utilized for each generation, the training pipeline for each model, and a discussion on subjective and quantitative evaluations of the models' Army-domain performance (see Section \ref{sec:Results}: Results \& Discussion). The training pipeline section is further broken down to capture distinguishing features of each generation, with special emphasis on techniques leading to improved performance over anterior generations.

\subsection{Training Data}

It is well-known among statisticians, data scientists, and machine learning specialists that data quality is paramount to creating a successful model. Furthermore, any model developed without sufficient data to accurately represent a domain or problem space will be inherently unreliable. Thus, from the onset of the TRACLM project, we sought a training corpus that was accessible, plentiful, and representative of the general Army domain. Army doctrine and related publications fit these criteria well. Due to the broad availability of Army publications over the open internet, we discovered that downloading and extracting the text from the unclassified subset of Army doctrine was near-trivial to automate with simple Python scripts. In March of 2023, and again in April of 2024, we accumulated over 4,300 unclassified documents from the Army Publishing Directorate (APD) website\cite{apd}. Many of these documents were hundreds of pages long,  resulting in an 80M+ token\footnote{NLP term representing individual words or sub-words, as defined by the target LLM's tokenizer. On average, one token equates to three-fourths of an English word, per popular tokenization algorithms\cite{tiktoken}.} corpus of high-quality Army-domain specific raw text (see Table \ref{tab:apd_corpus} in Section \ref{sec:Appendix}: Appendix for a full breakdown of TRACLM training corpus contents). For comparison, it is important to note that training LLMs in the 3-7B parameter range from scratch requires trillions of tokens for optimal convergence\cite{hoffmann2022training}. While the size of our APD corpus paled in comparison, our working assumption was that it represented the domain well enough that fine-tuning would prove effective. 

After downloading the documents from APD, we prioritized preprocessing before fine-tuning an LLM. Preprocessing is essential to cleanse the text of extraneous information and eliminate artifacts introduced by automated text extraction, such as extra whitespace and unnecessary symbols, which can impair the training process. Our preprocessing workflow evolved through successive TRACLM generations. Initially, we employed minimal preprocessing, followed by the integration of hard-coded rules to identify high-value data, and ultimately transformed the training corpus into synthetic questions and answers. Individual TRACLM generation preprocessing steps are documented at the beginning of each pipeline subsection.

\subsection{Training Pipeline}

The following subsections dissect TRACLM training pipelines into four distinct areas of discussion: data preprocessing, notable training hyperparameters\footnote{Training configurations under direct researcher control, e.g. number of training epochs. In contrast, model parameters are weight \& bias values that are updated automatically during training via backpropagation.}, hardware utilization, and miscellaneous notes. For TRACLM-v1, the Trainer class from the HuggingFace Transformers\cite{wolf-etal-2020-transformers} Python library was utilized for its effective abstraction of the PyTorch training loop\cite{Ansel_PyTorch_2_Faster_2024}. For training TRACLM-v2 and v3, we used the popular Axolotl library\cite{axolotl}, which provides an additional layer of abstraction over the HuggingFace Trainer that implements LLM training optimizations like sample packing\cite{krell2022efficient} and Microsoft's DeepSpeed\cite{deepspeed} framework for parallel training across GPUs. Unless otherwise noted, pipeline specifications from the initial generation apply to future generations.

\subsubsection{TRACLM-v1}

\begin{figure}[h!]
    \centering
    \includegraphics[width=0.75\textwidth]{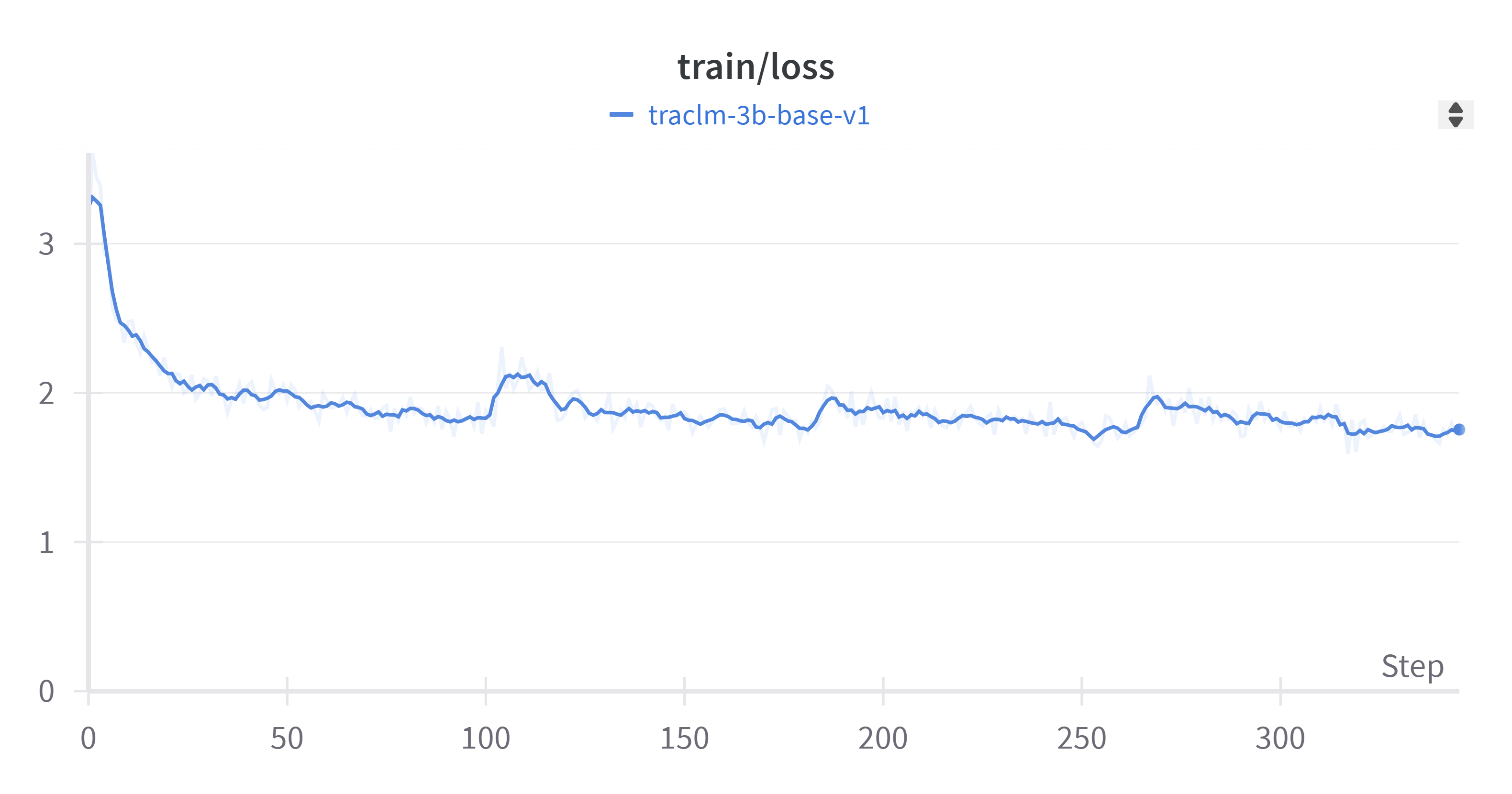}
    \caption{TRACLM-v1 Training Loss}
    \label{fig:traclm-v1}
\end{figure}

Data preprocessing for TRACLM-v1 was intentionally minimal, given our initial assumption that rigorous data cleaning was unnecessary given the base LLM's pretraining. Only two basic NLP rules were applied to our corpus before training: all text was converted to lowercase, and all leading or trailing continuous whitespace around sentences was converted to singular spaces. The base model we used for fine-tuning was Together Computer's RedPajama-INCITE-Base-3B-v1\cite{together2023incite}. This model was selected for its reputable source, permissive Apache 2.0 license, and comparable size to the open-source model we utilized in support of PC C4 in 2022. TRACLM-v1 was fine-tuned on our minimally preprocessed APD corpus for a single epoch using a batch size of one, 16 gradient accumulation steps, and an AdamW optimizer \cite{outmezguine2024decoupled}\footnote{Additional hyperparameters available at \url{https://huggingface.co/TRAC-MTRY/traclm-v1-3b-base}.}. The uninterrupted training run lasted 18 hours, and training was conducted in collaboration with the NPS Department of Defense Analysis for access to a 16x NVIDIA V100 GPU cluster. Figure \ref{fig:traclm-v1} illustrates TRACLM-v1's training loss\footnote{Quantification of the difference between LLM predictions and ground truth tokens, typically derived from categorical cross-entropy. Ideally, loss decreases while training as predictions improve over time.} throughout the course of fine-tuning. Notably, convergence occurred within the first quarter of the training run, suggesting a potential local minima was encountered early on.

\subsubsection{TRACLM-v2}

\begin{figure}[H]
    \centering
    \begin{minipage}[t]{0.48\textwidth}
        \centering
        \includegraphics[width=\textwidth]{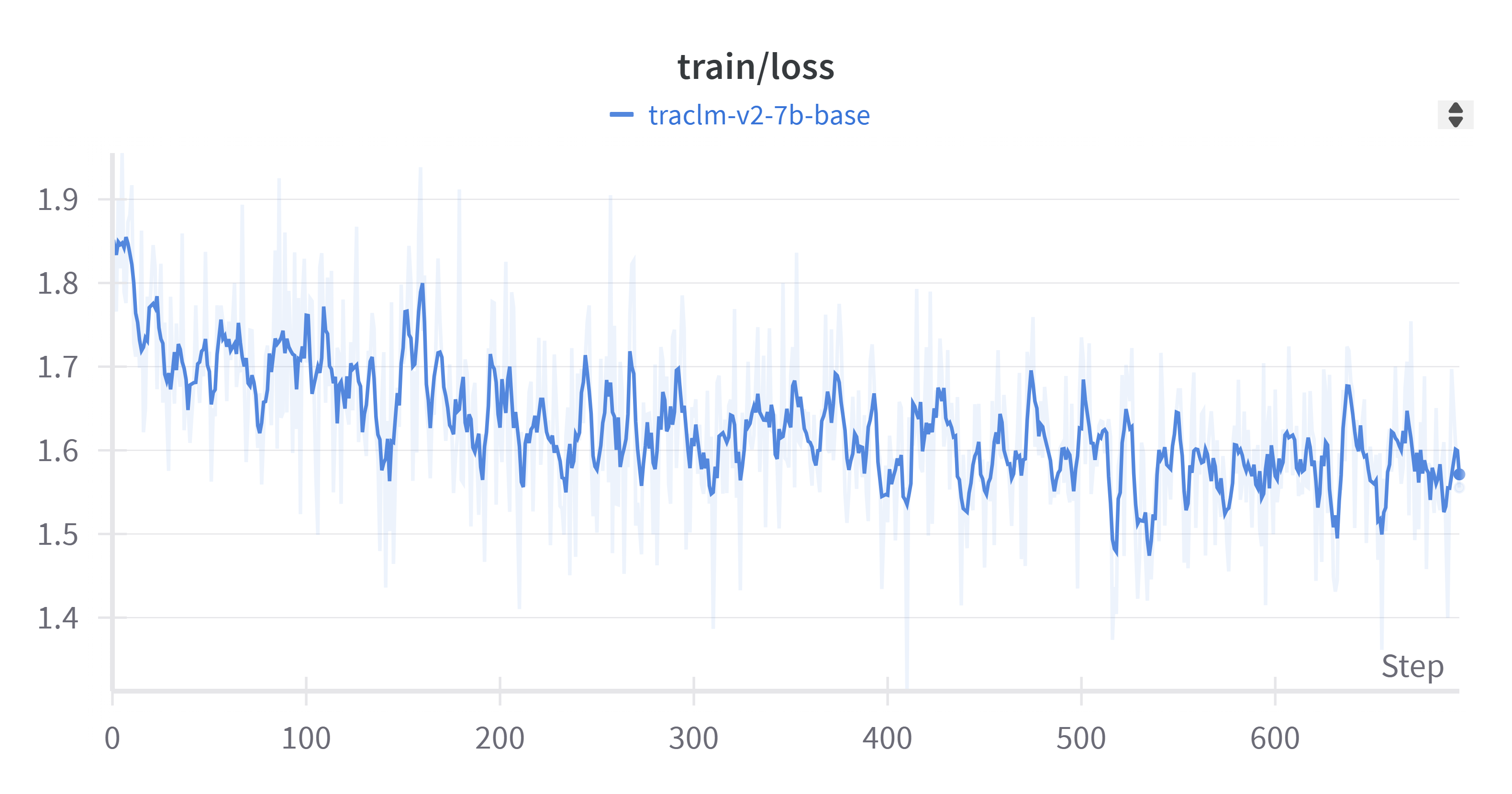}
        \captionsetup{justification=centering}
        \caption{TRACLM-v2 Training Loss\\(Base Model)}
        \label{fig:traclm-v2-base}
    \end{minipage}\hfill
    \begin{minipage}[t]{0.48\textwidth}
        \centering
        \includegraphics[width=\textwidth]{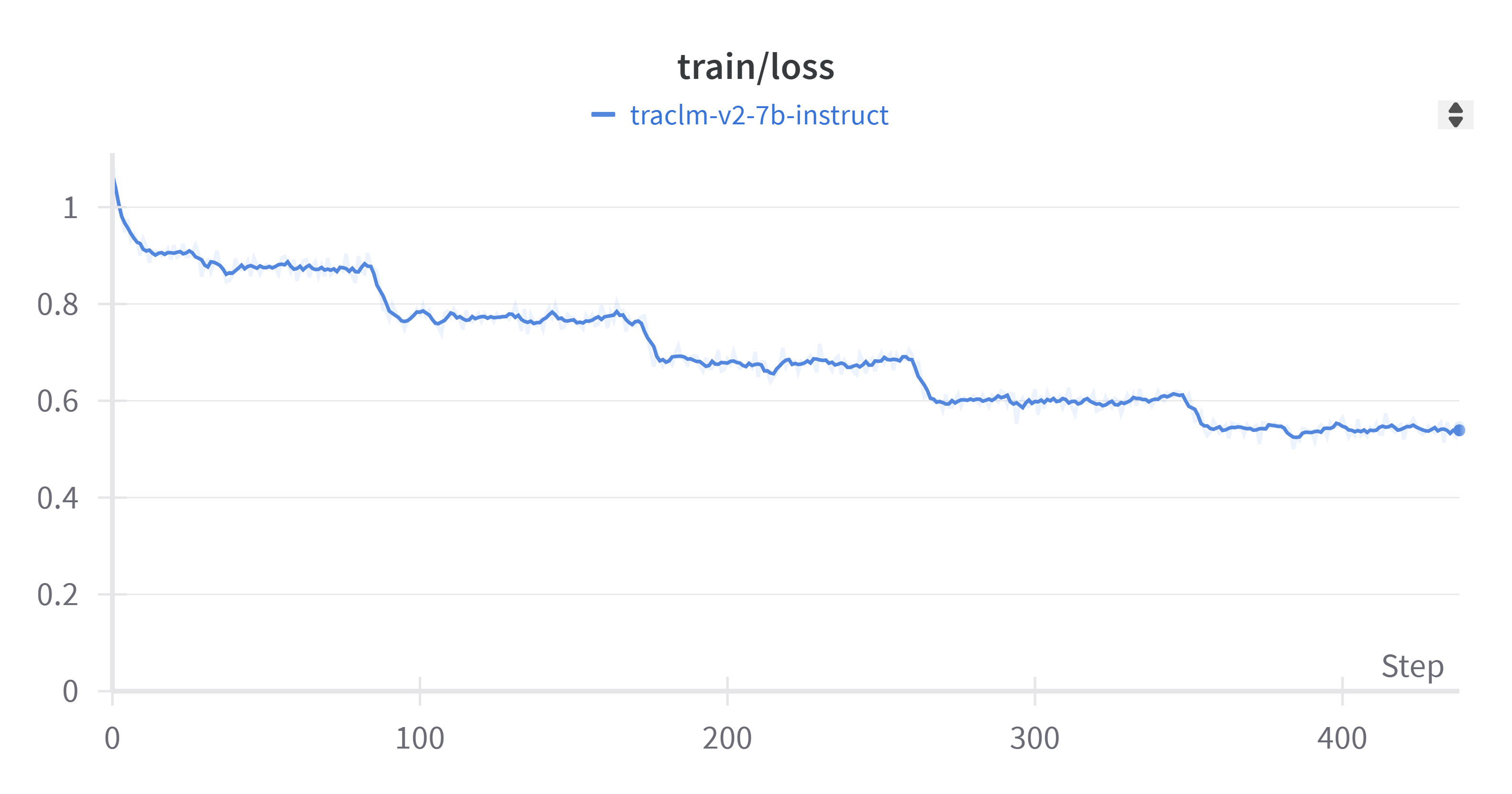}
        \captionsetup{justification=centering}
        \caption{TRACLM-v2 Training Loss\\(Instruction-tuned Model)}
        \label{fig:traclm-v2-instruct}
    \end{minipage}
\end{figure}

For TRACLM-v2, we implemented a number of changes to our training pipeline aimed at overcoming TRACLM-v1's performance shortfalls (see Section \ref{sec:Results}: Results \& Discussion). For data preprocessing, we incorporated additional basic NLP steps and custom rules to remove less relevant data from the corpus. Concretely, we excluded the tables of contents, references, and acknowledgments typically found at the beginning of documents, due to their sparse domain-specific knowledge and limited value for LLM training. These changes were made based on the observation that semantic relevance in our APD corpus varies across each document. We also upgraded our base model from RedPajama-INCITE-Base-3B-v1 to Meta's Llama-2-7b\cite{touvron2023llama}. This decision was made based on the general understanding that leveraging an increased parameter count from a more recently released open-source LLM would better tap into emergent capabilities\cite{wei2022emergent}. Lastly, perhaps the most impactful change to the TRACLM-v2 training pipeline was the addition of a second stage of fine-tuning after the continued pretraining stage. This "instruction-tuning" stage is meant to convert a raw LLM\footnote{LLM that has undergone requisite pretraining (typically, causal language modeling, also called next token prediction) and thus excels at continuing text but does not reliably follow instructions.} into a chatbot that responds well to questions and general instructions\cite{sanh2022multitask}. For instruction tuning, we used the open-source Alpaca dataset\cite{alpaca}, comprised of approximately 52,000 question \& answer pairs sourced from GPT-4. Both TRACLM-v2-7b-base and TRACLM-v2-7b-instruct (the instruction-tuned variant with second-stage fine-tuning) were fine-tuned for five epochs with a batch size of 4 and no gradient accumulation\footnote{Additional hyperparameters available at \url{https://huggingface.co/TRAC-MTRY/traclm-v2-7b-base} and \url{https://huggingface.co/TRAC-MTRY/traclm-v2-7b-instruct}.}. Training was conducted on an 8x NVIDIA A40 cluster maintained by the NPS High Performance Computing (HPC) Center. Each stage of training, namely continued pretraining and instruction-tuning, lasted 4.5 and 3.5 hours respectively, showcasing the advantages provided by upgrading from Volta to Ampere GPU architectures. Figures \ref{fig:traclm-v2-base} and \ref{fig:traclm-v2-instruct} illustrate TRACLM-v2's training loss during each stage of fine-tuning.

\subsubsection{TRACLM-v3}

Initially, we were satisfied with TRACLM-v2's domain-specific knowledge and instruction-following capabilities, but we soon realized it struggled with answering follow-up questions. We developed a working hypothesis that this capability shortfall was derived from two factors, namely, that our instruction-tuning dataset was not large enough and, more importantly, the instruction-tuning dataset did not contain any domain-specific examples. TRACLM-v3, our current flagship model, thus offers one critical innovation over version 2's training pipeline. To create a larger and more domain oriented instruction tuning dataset, we converted our APD corpus into approximately 500,000 question and answer pairs, and shuffled them together with an open-source dataset comprised of approximately 300,000 pairs. We leveraged synthetic example generation utilizing a larger, near-SOTA, open-source LLM: MistralAI's Mixtral-8x7B-Instruct-v0.1\cite{jiang2024mixtral}. In recent months, LLM training via synthetic examples has emerged as a promising approach in data-constrained circumstances\cite{liu2024best}; thus, we sought to employ related techniques for TRACLM-v3. Our abbreviated algorithm for question and answer generation via prompt engineering is represented below:

\begin{algorithm}
\caption{Q\&A Generation and Evaluation Workflow}
\begin{algorithmic}[1]
\STATE Split raw corpus into chunks
\FOR{each chunk}
    \FOR{each category}
        \STATE $attempts \gets 0$
        \REPEAT
            \STATE Prompt LLM for Q\&A within category
            \STATE Prompt LLM to evaluate Q\&A
            \STATE $attempts \gets attempts + 1$
        \UNTIL{quality threshold met \OR attempts $=$ 10}
        \IF{quality threshold not met}
            \STATE Label chunk as \texttt{<unsuitable for conversion>}
        \ENDIF
    \ENDFOR
\ENDFOR
\STATE Drop chunks labeled \texttt{<unsuitable for conversion>} from the final dataset
\end{algorithmic}
\end{algorithm}

\begin{figure}[H]
    \centering
    \includegraphics[width=0.75\textwidth]{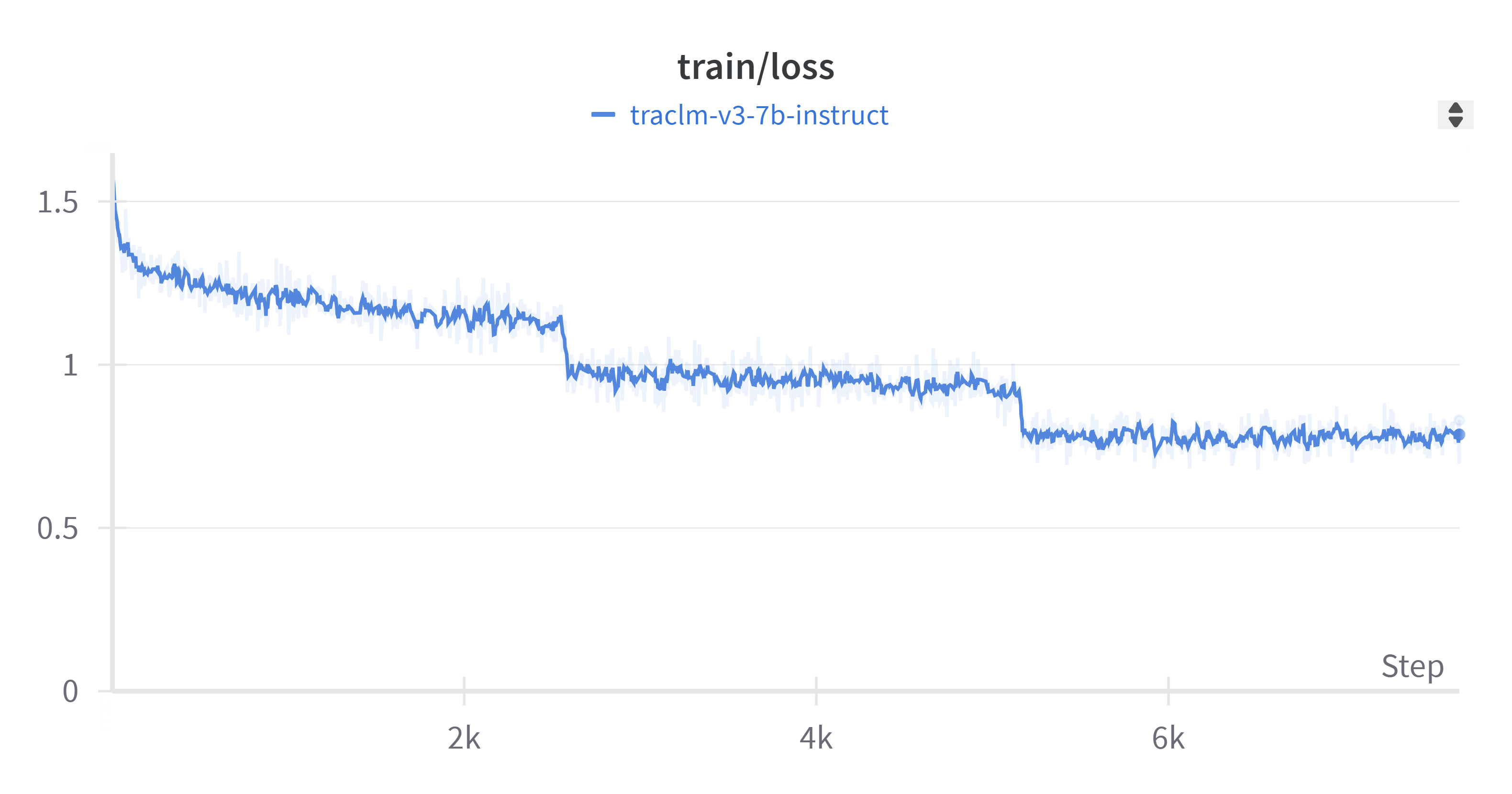}
    \caption{TRACLM-v3 Training Loss}
    \label{fig:traclm-v3}
\end{figure}

Question categories for generated examples were derived from recent Microsoft research exploring LLM domain adaptation\cite{cheng2024adapting}. Singleton examples of generated questions from each category are provided in Section \ref{sec:Appendix}: Appendix. Due to the resulting size of the TRACLM-v3 fine-tuning corpus, training was completed in a single stage in which an updated base LLM, MistralAI's Mistral-7B-v0.1\cite{jiang2023mistral}, was transformed directly into an Army-domain instruction-tuned model. This starkly contrasts TRACLM-v2's two-stage training pipeline, and further research is required to determine which approach is favorable under otherwise identical conditions (see Section \ref{sec:FutureWork}: Future Work). TRACLM-v3-7b-instruct was fine-tuned for three epochs with a batch size of 16 and no gradient accumulation\footnote{Additional hyperparameters: \url{https://huggingface.co/TRAC-MTRY/traclm-v3-7b-instruct}.}. Training was conducted on a 4x NVIDIA A100 workstation hosted within the NPS HPC center. Total training time for the final model was 26 hours, though if we factor in the time required to generate the synthetic dataset, time-to-train approaches 72 hours. Figure \ref{fig:traclm-v3} illustrates TRACLM-v3's loss value throughout the training run.

%% file: draft_content/2_milbench.tex
\section{MilBench}
\label{sec:MilBench}

Just as evaluation is essential to calibrate the accuracy and effectiveness of a weapon system, quantitative performance evaluation is essential for successful LLM fine-tuning. Throughout the first two fine-tuning iterations of TRACLM, we measured its performance qualitatively by observing the LLM's response to a set of standardized prompts. While this qualitative evaluation revealed an apparent improvement in performance from TRACLM-v1 to TRACLM-v2, it did not provide the detailed performance characterization necessary to understand how fine-tuning impacted TRACLM's performance across various domain-specific tasks. Furthermore, our previous evaluation method was also ill-suited to assess whether fine-tuning led to decreased performance on general tasks when compared to the base models, sans fine-tuning.

LLM evaluation has received significant academic and community interest, with HuggingFace's Open LLM Leaderboard\cite{open-llm-leaderboard} and EleutherAI's Language Model Evaluation Harness (LEH)\cite{llm-evaluation-harness} being notable examples of comprehensive performance leaderboards and evaluation frameworks, respectively. Given that LEH is open-source with an MIT license, we considered leveraging it to evaluate TRACLM. However, we found its complexities went beyond our use case and opted to build our own framework, called MilBench, that leveraged an open-source dataset format\footnote{Both MilBench and LEH are compatible with the Hugging Face Datasets format (\url{https://huggingface.co/docs/datasets/}).} such that our tasks would be interoperable with LEH, and our framework would be compatible with community-made datasets.

MilBench is a collection of benchmarking datasets aggregated and maintained by TRAC and a modular software framework that enables organizations of all sizes to easily evaluate and assess LLMs at scale. In this section, we describe each of MilBench's three components: MilBench Datasets, MilBench Evaluation Harness (MEH), and MilBench Server. We highlight how MilBench can be an invaluable part of not only the DoD LLM research community, but also the Generative AI acquisitions process, providing organizations a user-friendly web interface to start, manage, compare, and audit quantitative LLM evaluations.

\subsection{MilBench Datasets}

To quantitatively evaluate TRACLM and other LLMs, we sought a methodology and datasets that provided meaningful performance metrics and LLM compatibility. While there exist many methods of evaluating LLMs, several of the widely-used benchmarks are comprised of multiple-choice questions and answers\footnote{MMLU\cite{hendrycks2021measuring}, TruthfulQA\cite{lin2022truthfulqa}, and ARC\cite{clark2018think} are some of the most popular multiple-choice evaluation datasets at the time of writing.} that are used to prompt a model with a question and ask it to respond with the letter of the correct response. Using multiple-choice questions provides meaningful performance metrics as we can compare the scores of an LLM to that of a human taking the same test. Multiple-choice questions are also faster and cheaper to evaluate compared to questions that require longer responses, as the LLM usually needs to generate only one token to provide a ready-to-evaluate response.

MilBench Datasets is a repository of datasets, referred to as tasks, that are designed to evaluate LLM performance in the military domain. The current repository contains four tasks derived from those presented by Hallapy et al. in MilGLUE\cite{milglue} and one task derived from Army officer multiple-choice tests (CATB). As mentioned above, MilBench supports any multiple-choice dataset in the HuggingFace Datasets format, to include common benchmarks like MMLU\cite{hendrycks2021measuring}. Though we will limit further discussion of non-Army datasets, we emphasize that this standard format enables any organization to create their own evaluation datasets for use within MilBench.

\subsubsection{MilGLUE-Derived Tasks}

MilGLUE \cite{milglue} presents evaluation datasets that are tailored for Bidirectional Encoder Representations from Transformers (BERT) \cite{bert} models. Thus, raw MilGLUE examples are neither conversational in nature nor formatted as multiple-choice questions. However, this does not present a problem for our use case as it is common practice when evaluating LLMs to wrap the dataset entries with conversational cues as part of a task definition within the evaluation framework. The MilGLUE dataset corpus required further curation due to its size, nearly 1,000,000 records for some datasets, which we deemed too expansive for large-scale LLM evaluation due to limited compute and time resources. We selected four of the MilGLUE datasets to curate for use with MilBench, as shown in Table \ref{tab:milbench-datasets} and denoted in bold.

\begin{table}[H]
  \caption{MilBench Datasets}
  \label{tab:milbench-datasets}
  \centering
  \begin{tabular}{p{0.4\linewidth}p{0.4\linewidth}r}
    \toprule
    Task Name     & Description     & Size (Records) \\
    \midrule

    \textbf{Masked Reasoning (MR)}     &
    Choose the word that correctly replaces a masked word in a sentence. &
    10,000   \\

    \textbf{Next-Sentence Reasoning (NSR)}     &
    Determine whether a sentence correctly follows another sentence. &
    10,000  \\

    \textbf{Paraphrase (PARA)}     &
    Determine whether a sentence correctly paraphrases a paragraph. &
    10,000 \\

    \textbf{Sentence Similarity Binary (SSB)}     &
    Determine whether two sentences are semantically identical. &
    10,000  \\  

    Combined Army Test Bank (CATB) &
    Collection of questions from Army professional military education and U.S. Military Academy courses.  &
    577  \\
    \bottomrule
  \end{tabular}
\end{table}

Our curation methodology aimed to reduce the time required to evaluate an LLM by selecting a subset of the records from each dataset that were semantically representative of the whole. We selected 10,000 records from each dataset using a combination of rules-based filtering and topic modeling. All records that contained "blanks", strings of ten or more underscores, were removed. Records that were not entirely made up of ASCII characters were also removed. 

To ensure that sampled records were semantically representative of the original dataset, we then embedded each remaining record into a high-dimensional vector space using all-mpnet-base-v2, the best-performing model for this task at the time \cite{sentence-transformers} \cite{sentence-transformers-paper}. BERTopic \cite{bertopic} was then used to group the embedded vectors representing our records, seeded with the value $1234$. We selected eight sets of keywords as seed topics (See Table \ref{tab:milglue_topic_seeds}). We then used BERTopic's reduce\_topics method to reduce the total number of discovered topics to no more than twenty via Agglomerative Clustering. The keywords representing each topic and the size of each topic are listed in the Appendix (Section \ref{sec:MilGLUE_Dataset_Makeup}: MilGLUE Dataset Make-Up). 
We then selected 10,000 records from each topic using a round-robin strategy, as shown in Algorithm \ref{alg:record_selection}. This ensured our subset topic proportions matched those of the original MilGLUE datasets, greatly increased our confidence in using them for evaluation.

\begin{algorithm}
\caption{Record Selection Algorithm (Single Dataset)}\label{alg:record_selection}
\begin{algorithmic}[1]
\STATE $topic\_records\_map \gets \text{Map of topics to record sets, sorted by topic ID}$

\STATE $selected\_records \gets \emptyset$

\WHILE{$|selected\_records| < 10,000$}

    \FOR{topic, records in $topic\_records\_map$}
        \WHILE{$records$ is not empty}
            \STATE $record \gets \text{pop a record from }records$

            \IF{$record$ is clean}
                \STATE $selected\_records \gets selected\_records \cup \{(topic, record)\}$
                \STATE \textbf{break}
            \ENDIF

            \IF{$|selected\_document\_ids| == set\_size$}
                \STATE \textbf{break}
            \ENDIF
        \ENDWHILE

        \IF{$|selected\_document\_ids| == set\_size$}
            \STATE \textbf{break}
        \ENDIF
    \ENDFOR
\ENDWHILE

\end{algorithmic}
\end{algorithm}

\subsubsection{CATB Task}

The CATB task is of special mention, as it enables a user-friendly form of evaluation that is easily understood by virtually all uniformed service members. The intent for CATB is to aggregate Army multiple-choice questions, starting with the United States Military Academy (USMA) military science program and continuing through increasingly advanced Army professional military education (PME). Upon evaluating an LLM with the CATB task, it will become possible to "rank" LLMs based on their performance across this broad range of exams. To date, we have collected USMA Department of Military Instruction (DMI) questions through official releases, and Army Command and General Staff College (CGSC) questions from online study guides containing outdated test questions\cite{quizlet_ile}. Future agreements, which are in progress at the time of writing, will add content from the Army's Basic Officer Leader Course (BOLC), Captain's Career Course (CCC), and other specialized PME sources. In its current form, despite lacking diverse sets of questions from various sources, the CATB task has still proven effective for informing our fine-tuning process as a means of validating the improved domain-specific knowledge of each subsequent model version.

\subsection{MilBench Evaluation Harness}

MEH is a modular evaluation framework written in Python that serves as a test proctor for LLMs hosted on a HuggingFace Text Generation Inference (TGI) server, or behind any OpenAI-compatible application programming interface (API).
We chose to have MilBench offload interaction with LLMs to inference servers as this enables MEH to concurrently evaluate a greater number of models than its host machine could run simultaneous inference on.
The harness can be configured to administer an arbitrary set of tasks, so long as the associated datasets are in the HuggingFace Datasets format, a format used by all common LLM evaluation datasets.
As of writing, MilBench exclusively supports evaluation using tasks comprised of multiple-choice questions. These tasks are defined using a YAML format inspired by that of LEH\footnote{See Section \ref{sec:MEH_Configuration} for
the MMLU task configuration.}. The task configuration format used by MilBench notably omits parameters such as $group$, $metric\_list$, and $generation\_kwargs$ when compared to LEH. These omissions are a direct result
of the intentional simplicity of MEH, and do not present significant friction when porting a task definition from LEH to MEH.\footnote{The amount of friction involved in porting a task from LEH to MEH is an important consideration, as lower friction enables more tasks built for LEH to be used with MEH. Evaluating LLMs on more tasks gives a clearer picture of their capabilities across various subjects and domains.}

\begin{figure}[H]
    \centering
    \includegraphics[width=0.75\textwidth]{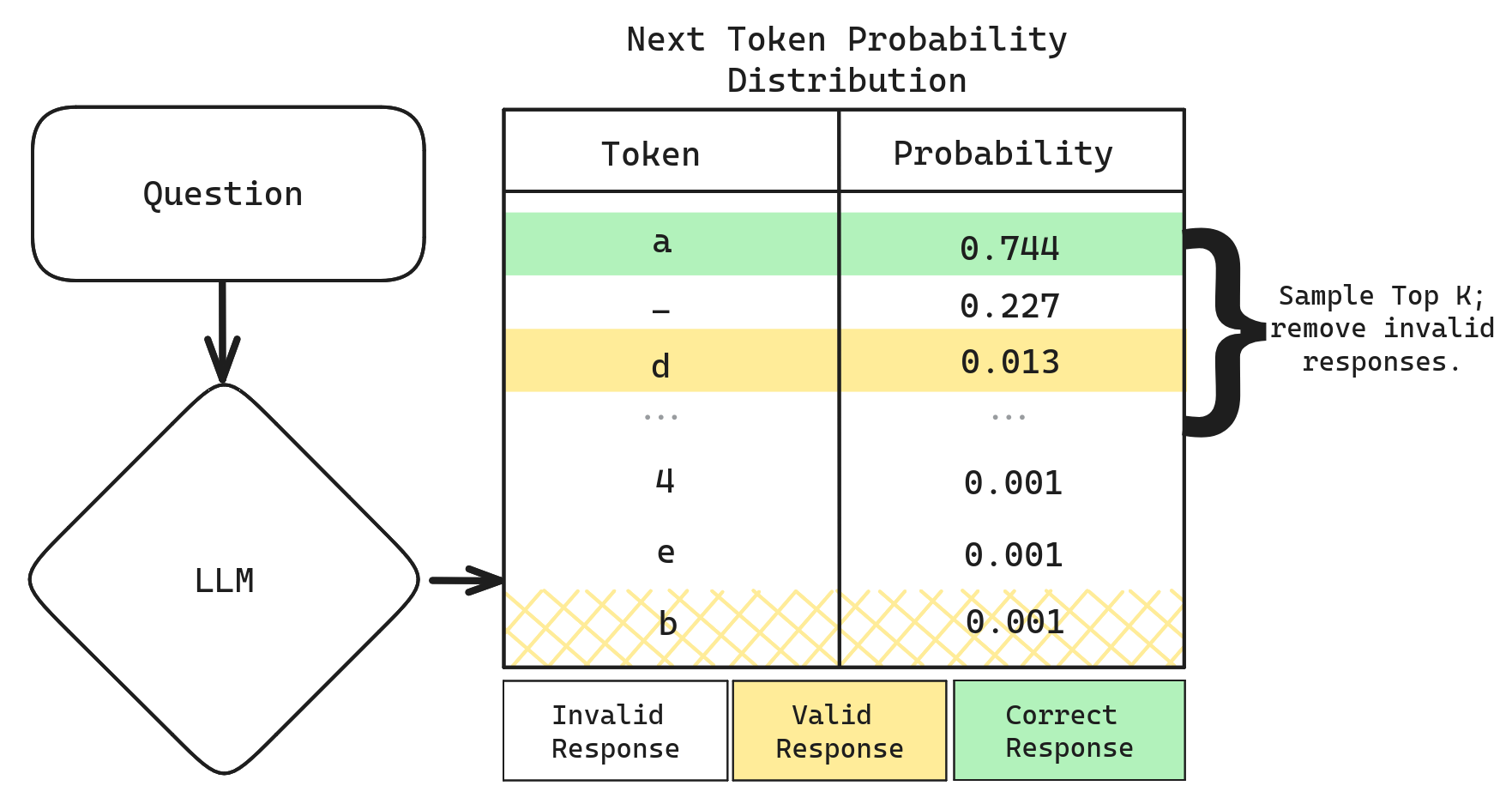}
   \caption{MilBench Evaluation Metric}
    \small Note: In practice, only the top k tokens are ever considered. The entire distribution is shown here for illustrative purposes.
    \label{fig:milbench_evaluation}
\end{figure}

Because LLMs are fundamentally prediction machines that produce as output a probability distribution (PD) of tokens that follow the input, there are many strategies to decide which token from the distribution is to be considered the LLM's response. The simplest strategy is to greedily choose the token with the highest probability. Another strategy is to randomly choose from the top $k$ most probable tokens, known as $top\_k\ sampling$. Sampling, however, is not always the best choice when performing LLM evaluation. For example, it may be desired (and expected) that an LLM chatbot would respond with slight phrasing differences to a multiple choice question. When prompted with the same question, it may respond directly with a letter containing the response, or it may begin its response with "The correct answer is". In order to reduce computational resources required for evaluation, we desired to only run a single forward pass (generate one probability distribution) for each question. In consideration of the previously mentioned behavior, the strategy that we employ within MEH is a modified version of $top\_k\ sampling$. First, we take the top five most probable tokens from the PD. We then remove all tokens that, when normalized by lower-casing and white-space trimming, are not valid responses to the question. Finally, we consider the most probable valid response from the top 5 tokens to be the model's response to the question (see Figure \ref{fig:milbench_evaluation}). This technique enables us to evaluate models that might tend to prefix their response with a phrase, while not being too lenient in our evaluation\footnote{One such lenient strategy would consider the LLM to have responded correctly if the correct response is found anywhere within a top $p$ probability range.}.

\subsection{MilBench Server}

To add to the functionality and utility of MEH, we then targeted capabilities beyond those enabled by a command line interface (CLI). These included easy initiation of evaluations from a remote computer, and providing a way to track, compare, and audit evaluation results. MilBench Server, a web API wrapper around MEH, was created to enable remote and distributed management of MEH evaluations. MilBench Server provides an easy-to-use RESTful\footnote{Choosing to have MilBench's communication adhere to a REpresentational State Transfer (REST) architecture enables any application to interact with it programatically, so long as it can consume (JSON) JavaScript Object Notation data.} API to start, view, and audit LLM evaluations. MilBench Server stores logs of every evaluation to enable leaderboard visualizations, as well as question-level visibility into evaluation results.
Moreover, we designed a web-based graphical user interface to streamline the process of initiating and monitoring evaluations. The interface is designed to be intuitive and user-friendly. In addition to an evaluation management dashboard, it includes a leaderboard and radar chart (as depicted in Figure \ref{fig:milbench_leaderboard}) that facilitates the visualization of the performance attributes of the assessed models.

Transparency in evaluation is, and will continue to be, key to understanding and building confidence in benchmark datasets and evaluation frameworks. MilBench Server, with its user-interface, presents a no-code view inside each evaluation at question-level resolution (See Figure \ref{fig:milbench_audit}). This level of detail is critical to understanding both high- and low-scoring LLMs, and is particularly helpful for teams aiming to increase LLM performance by fine-tuning as anomalous outputs can be visually identified. These anomalies may point to issues with training data, or perhaps tokenizer configuration errors. While objective quantitative evaluation is integral to understanding LLM performance, subjectively evaluating LLMs with a chat conversation is also critical --- especially for LLMs that are intended to be user-facing. To that end, the MilBench Server's user-interface supports simultaneous side-by-side chat conversations with up to ten LLMs (as shown in Figure \ref{fig:milbench_chat_evaluation}).

MilBench Server supports fine-tuning and acquisition workflows from start to finish. Fine-tuning efforts can leverage the existing leaderboard of LLM performance to identify baseline expectations. Once fine-tuning has been completed, the LLM can be quickly evaluated and added to the leaderboard for rapid comparison. This cycle can be repeated for subsequent fine-tuning efforts to visualize the effects of fine-tuning with respect to both domain adaptation and loss of general knowledge. Further, the broad use of MilBench and population of its leaderboard provide valuable context and enable data-driven decision making for Generative-AI acquisition decisions. Candidate models can be evaluated and compared to fine-tuned models using quantitative metrics to inform whether the candidate is worth acquiring.

%% file: draft_content/3_results.tex
\section{Results \& Discussion}
\label{sec:Results}

In this section, we report and discuss the performance of TRACLM models, both against each other and comparable open-source models in the 3-7B parameter range. We begin by quantifying the domain-specific knowledge of each generation of TRACLM, to demonstrate a progressive growth in knowledge as our data preprocessing and training pipeline evolved. Then, we compare the capabilities of our latest and most performant model, TRACLM-v3, against competing models via both quantitative and qualitative benchmarking.

\subsection{Comparing TRACLM Generations}

With each successive version of TRACLM, our prevailing goal was to fine-tune models that clearly outperform the preceding version in terms of domain-specific knowledge. While this difference in capability can be observed through basic interaction with each model, the reliability of this form of subjective analysis plateaus as models become more performant. Therefore, a tool enabling quantifiable comparison of LLMs against concrete and domain-specific benchmarks became essential to measure the increasing innate knowledge of our models. As discussed in Section \ref{sec:MilBench}, MilBench was developed to fulfil this critical role. For the purpose of our analysis here, we provide the following table and corresponding radar plot. Note the tasks labeled CATB through SSB are Army-specific benchmarks, while MMLU\cite{hendrycks2021measuring} and TQA\cite{lin2022truthfulqa} are open-source benchmarks meant to assess LLM general knowledge and truthfulness, respectively. The latter two benchmarks are included by default in MilBench to enable researchers to identify shortfalls in general capabilities while calibrating models for specialized domains.

\begin{table}[h!]
\centering
\caption{Performance across TRACLM Versions (\textsuperscript{\textdagger}denotes Army-specific task)}
\label{tab:traclm-performance-internal}
\begin{tabular}{lrrrrrrrr}
\toprule
\multicolumn{1}{c}{MODEL} & \multicolumn{1}{c}{Average} & \multicolumn{1}{c}{CATB\textsuperscript{\textdagger}} & \multicolumn{1}{c}{MR\textsuperscript{\textdagger}} & \multicolumn{1}{c}{NSR\textsuperscript{\textdagger}} & \multicolumn{1}{c}{PARA\textsuperscript{\textdagger}} & \multicolumn{1}{c}{SSB\textsuperscript{\textdagger}} & \multicolumn{1}{c}{MMLU} & \multicolumn{1}{c}{TQA} \\ \midrule
traclm-v3-7b-instruct & \textbf{69.20} & \textbf{69.93} & \textbf{78.21} & \textbf{85.63} & 65.65 & 73.33 & \textbf{61.33} & \textbf{50.29} \\
traclm-v2-7b-instruct & 52.19 & 57.69 & 51.22 & 65.26 & 39.87 & \textbf{73.61} & 44.62 & 33.04 \\
traclm-v1-3b-base & 36.11 & 31.47 & 50.47 & 36.26 & \textbf{67.06} & 23.21 & 24.24 & 20.03 \\ \bottomrule
\end{tabular}
\end{table}

As evidenced from Table \ref{tab:traclm-performance-internal}, each successive generation of TRACLM improved markedly over the preceding version(s), both in average performance and almost all individual tasks. These results are promising, despite the fact MilBench tasks have yet to be finalized. Further investigation is required to determine the anomalously high performance of TRACLM-v1 on the paraphrase identification (PARA) task. The following radar plot is also provided for an alternative visual representation of domain knowledge acquisition:

\begin{figure}[h!]
    \centering
    \includegraphics[width=0.75\textwidth]{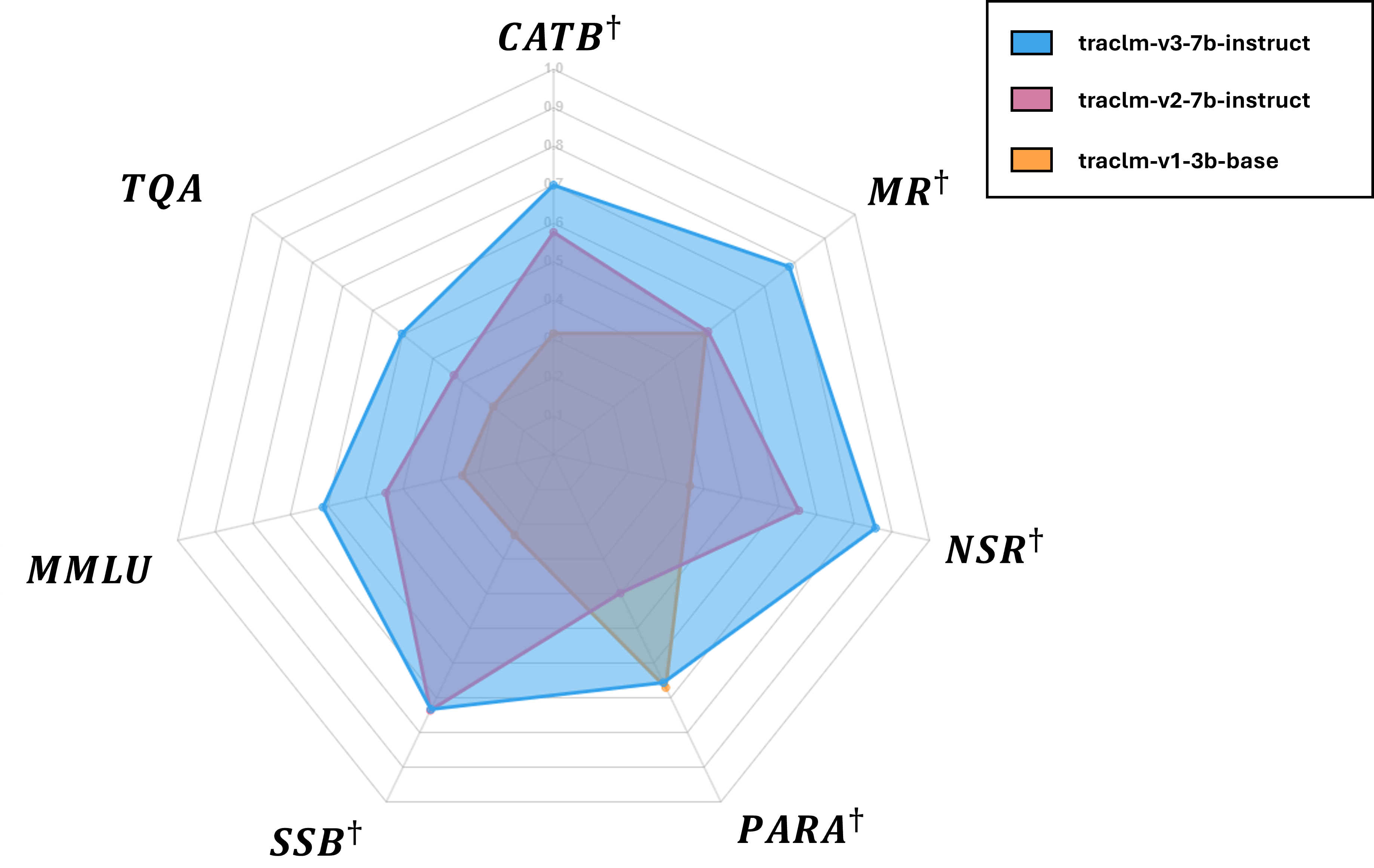}
    \caption{TRACLM Version Comparison on MilBench Tasks (\textsuperscript{\textdagger}denotes Army-specific task)}
    \label{fig:traclm-comparison}
\end{figure}

\subsection{TRACLM-v3 Performance Analysis}

TRACLM-v3's domain adaptation is evident when contrasted against well-known open-source models, both quantitatively and qualitatively. Table \ref{tab:traclm-performance-external} uses MilBench to compare TRACLM-v3 against three instruction-tuned variants of popular 7B-parameter LLMs, and one mixture-of-experts LLM with 56B parameters. Despite the 8x increase in parameter count of Mixtral-8x7B-Instruct-v0.1, TRACLM-v3 still manages to outperform it on 3/5 of the domain-specific tasks, heavily supporting the efficacy of the TRACLM-v3 training corpus and pipeline.

\begin{table}[h!]
\centering
\caption{Performance across Competing Models (\textsuperscript{\textdagger}denotes Army-specific task)}
\label{tab:traclm-performance-external}
\begin{tabular}{lrrrrrrrr}
\toprule
\multicolumn{1}{c}{MODEL} & \multicolumn{1}{c}{AVG} & \multicolumn{1}{c}{CATB\textsuperscript{\textdagger}} & \multicolumn{1}{c}{MR\textsuperscript{\textdagger}} & \multicolumn{1}{c}{NSR\textsuperscript{\textdagger}} & \multicolumn{1}{c}{PARA\textsuperscript{\textdagger}} & \multicolumn{1}{c}{SSB\textsuperscript{\textdagger}} & \multicolumn{1}{c}{MMLU} & \multicolumn{1}{c}{TQA} \\ \midrule
traclm-v3-7b-instruct & \textbf{69.20} & 69.93 & \textbf{78.21} & \textbf{85.63} & 65.65 & 73.33 & 61.33 & 50.29 \\
mistral-7b-instruct-v0.1 & 62.69 & 60.84 & 66.24 & 75.12 & 62.38 & 73.71 & 52.02 & 48.54 \\
mixtral-8x7b-instruct-v0.1 & 66.02 & 70.28 & 76.57 & 57.91 & 43.81 & \textbf{73.90} & \textbf{68.10} & \textbf{71.60} \\
meta-llama-3-8b-instruct & 63.06 & \textbf{72.90} & 75.11 & 59.45 & \textbf{67.54} & 63.36 & 66.64 & 36.40 \\
zephyr-7b-beta & 58.50 & 62.59 & 67.60 & 52.20 & 47.32 & 73.81 & 57.29 & 48.68 \\
\bottomrule
\end{tabular}
\end{table}

\begin{figure}[H]
    \centering
    \includegraphics[width=0.75\textwidth]{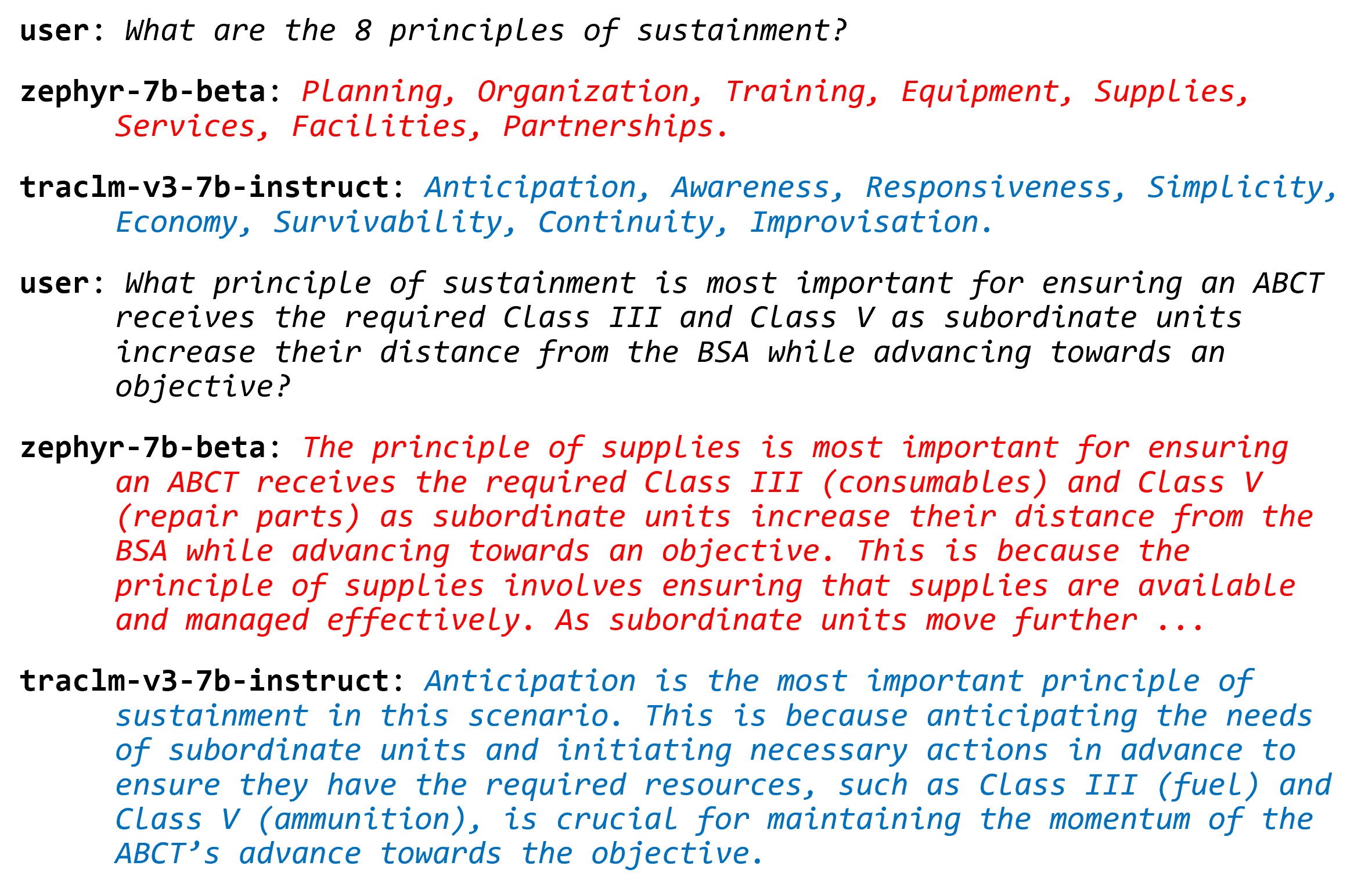}
    \caption{Zephyr-7b-beta vs. TRACLM-v3 Response Quality, Exhibit A}
    \label{fig:sustainment_example}
\end{figure}

To demonstrate the domain-specific knowledge of our best model, Figure \ref{fig:sustainment_example} provides a transcript of a simultaneous interaction  between TRACLM-v3 and Zephyr-7b-beta\cite{tunstall2023zephyr}, a competing Mistral-7B-v0.1 fine-tune that underwent both instruction-tuning and helpfulness alignment via DPO. Both models are provided the same query by the user, and the truncated responses from both models are provided in red and blue for comparison. A follow-up question utilizing multiple Army terms and acronyms is then provided, and the model responses are once again captured for comparison.

In this example, TRACLM-v3 not only provides the principles of sustainment as stated in U.S. Army Doctrine with perfect recall, it also demonstrates a deeper understanding of associated concepts by selecting a sensible principle for the follow-up question and correctly unpacking the stated classes of supply\cite{FM4-0}, all of which the competing model is unable to do. The reasons for this disparity are clear: Zephyr-7b-beta, while highly performant and instruction-tuned, can only provide answers that approximate the content its been trained on, i.e. general-purpose internet-sourced text. TRACLM-v3, on the other hand, has been calibrated through fine-tuning such that its responses are better suited for Army-domain usage. Additional comparisons representing other Army sub-domains are provided in Section \ref{sec:Appendix}: Appendix.

%% file: draft_content/4_winddown.tex
\section{Conclusion}

In this paper, we introduced TRACLM, a first-of-its-kind series of LLMs fine-tuned on unclassified Army doctrine for general application in the Army domain. We shared our methodology and techniques, assumptions, analysis, results, and best practices on military domain LLM calibration based on our experiences for the benefit of the Army analytical community. Additionally, we introduced MilBench, a modular and model-agnostic evaluation framework leveraging custom Army-specific benchmarks to assess any open-source, locally-hosted LLMs. MilBench possesses high impact potential across the DoD, because it enables any government organization to incorporate quantitative LLM evaluations into their generative AI acquisition pipelines, thus enhancing senior leaders' ability to make informed decisions about commercial partnerships vs. in-house development.

\section{Acknowledgments}

Due to the heavy computational requirements involved in training LLMs, this project would not have been possible without the gracious support of the NPS Department of Defense Analysis and NPS HPC Center. Furthermore, MIT-LL's generous sharing of benchmark data from the MilGLUE project served as an invaluable starting point for MilBench tasks. Furthermore, we acknowledge the USMA Department of Military Science's support for the MilBench project, demonstrated by the official release of test questions for inclusion in the CATB task. Lastly, we thank the interns, analysts, and leaders currently assigned to TRAC for their assistance in refining this paper. Special mention is given to Ms. Sarah Harkins, MAJ Ryan Davis, Dr. Brian Wade, and LTC James Jablonski.

\section{Limitations}

As a snapshot of a continuous research project, TRACLM-v3 suffers from some limitations despite the capabilities demonstrated throughout this paper. Firstly, TRACLM-v3's context window mirrors that of the base model: 4096 tokens. Effectively, this means model performance will swiftly diminish when asked to process sequences longer than this maximum, which renders certain use cases (e.g. long transcript summarization) intractable until future versions. Second, and closely related, we have observed through qualitative evaluations that TRACLM-v3's rate of hallucination increases as the context limit is approached. We theorize this effect is influenced not only by the limited context window, but also by the often direct and to-the-point tendencies of Army communications, suggesting that TRACLM-v3 "prefers" to arrive at its main points and terminate generation as soon as possible. Third, TRACLM-v3 is prone to false attribution regarding its creators and affiliation. Concretely, when prompted with certain verbiage, TRACLM-v3 will confidently state it is an "LLM created by OpenAI". This is due to the presence of such statements in the open-source subset of the TRACLM-v3 training corpus that were not removed prior to training. Lastly, the aforementioned limitations of TRACLM-v3 are amplified for TRACLM-v2, due to the less sophisticated nature of the latter's training pipeline. Also of note, TRACLM-v1 is severely limited by its classification as a raw LLM, meaning it is incapable of following any instructions reliably.

MilBench suffers from similar issues to other benchmarks and evaluation frameworks. The CATB dataset has been observed to have a small portion of its questions reference contextual information that would have been present in the test's original form, but is not present in the benchmark dataset. Additionally, CATB includes questions that use strings of underscores that vary in length to represent "fill-in-the-blank" questions. We theorize that such inconsistent and implicit question formatting may introduce uncertainty into LLM response correctness. Records in the MilGLUE tasks have been noted to contain data that is either grammatically flawed, or that suffers from formatting issues. For example, the paraphrase task has at least one record where the source paragraph contains a full URL, but the proposed paraphrase has a malformed URL where all special characters are removed. Such dataset issues introduce some amount of error into evaluation results, as missing context or malformed text may lead to multiple responses being plausible. MEH is further limited by only supporting the filtered $top\_k$ evaluation strategy, which is only applicable to questions whose response can be expected to be a single token. Due to these limitations, the MilBench tasks are not yet finalized as there remains work to further curate and clean the datasets. While this does mean that an LLM's score on a given task is subject to change, we do not anticipate the datasets changing enough to invalidate our preliminary results. Additionally, the initial findings will continue to provide valuable insight to the fine-tuning process.

\section{Safety \& Security Considerations}

Despite TRACLM-v3's clear ability to generate accurate, Army-specific text, it is still a probabilistic model, and thus is capable of producing false information with little to no effort by the user. Thus, researchers employing TRACLM for their own experiments should verify all outputs. The creation of the TRACLM series of models constitutes academic research in partnership with the Naval Postgraduate School. The purpose of this research is to inform future DoD experimentation regarding the development and application of domain-specific LLMs. While experiments involving direct application of this model to downstream military tasks are encouraged, extreme caution should be used before production-level integration into Army systems. Regarding classification risks, TRACLM is trained on exclusively unclassified and publicly released training data. Therefore, we deem the risk of TRACLM producing higher classification content very unlikely. However, classification by compilation is a risk associated with all LLMs that is of unique concern to the DoD. Thus, we invite researchers from across the analytical community to leverage TRACLM in their future studies to help quantify LLM classification risks with concrete findings.

Although we intend for MilBench to be leveraged across the DoD to evaluate LLMs, it is important that the evaluation datasets remain close-hold. Exposure of the raw datasets would enable LLMs to be trained on the exams themselves, which would lead to greater performance scores but would render those scores meaningless as evaluations are only valid if the models have not previously seen the questions and answers. This may theoretically prevent MilBench from evaluating commercial LLMs unless it could be guaranteed that no inference logs were persisted during evaluation.  Additionally, we acknowledge that any leaderboards showing the performance of DoD fine-tuned LLMs are best maintained internally. Unrestricted access to such leaderboards would risk advertising to U.S. adversaries which models possess the highest amount of domain-specific knowledge. Therefore, there are no plans to host MilBench leaderboards on public-facing domains like the HuggingFace Open LLM Leaderboard.\cite{open-llm-leaderboard}
\section{Future Work}
\label{sec:FutureWork}

Future work associated with the TRACLM project aims to produce more capable model versions using larger base models with longer context windows. Furthermore, we plan to conduct extensive experiments to conclusively inform the preceding discussions regarding two-stage (e.g. continued pretraining followed by instruction-tuning) vs. single-stage (e.g. instruction-tuning only) fine-tuning. We will also continue to improve the TRACLM training corpus by including data from additional sources (e.g. other Army PME courses and authoritative documents maintained by each of the Army's Centers of Excellence) and address aforementioned attribution issues with open-source instruction-tuning datasets. Lastly, we'll explore the value of model alignment via newer techniques like DPO, which will require the creation of additional Army-specific training examples derived from recorded interactions with prior TRACLM versions.

Future work on MilBench includes the collection of more domain-specific evaluations from Army officer PME, Army enlisted basic and advanced individual training (AIT), and Army enlisted PME such that LLM performance could be loosely categorized into existing tiers of service member performance\footnote{Gathering datasets from evaluations that take place throughout an Army career enables comparisons such as "LLM X performs at a level roughly equivalent to that of an average Army Captain."}. We also plan to further sanitize and finalize our domain-specific benchmarks. Continued work on MEH aims to add support for evaluation strategies that can consider more than a single token response, and we plan to add support for evaluating a model's performance on a specific task, such as retrieval-augmented-generation. Finally, we plan to add robust multi-user support to MilBench Server to enable broader access to its evaluation capability. 

%% file: draft_content/5_appendix.tex
\section{Appendix}
\label{sec:Appendix}

\subsection{TRACLM Training Data Breakdown}

Table \ref{tab:apd_corpus} details the contents of the raw TRACLM training corpus, as extracted in April 2024. Collections of documents are delineated according to their type, with viable token counts and percentages provided per type. Viable tokens are those included in the finalized corpus for TRACLM-v2 (i.e. tokens not removed via preprocessing).

\begin{table}[H]
\centering
\caption{APD Corpus Breakdown, as of April 2024.}
\label{tab:apd_corpus}
\begin{tabular}{@{} llll l @{}}
\toprule
\textbf{Type} & \textbf{Description} & \textbf{Doc Count} & \textbf{Viable Tokens} & \textbf{\% Total} \\ \midrule
AR      & Army Regulations                & 470  & 16425075 & 20\% \\
ATP     & Army Technical Publications     & 191  & 15500853 & 19\% \\
TM      & Technical Manuals               & 166  & 12308025 & 15\% \\
TC      & Training Circulars              & 102  & 8657219  & 10\% \\
AGO     & Army General Orders             & 2901 & 8071842  & 10\% \\
PAM     & Department of the Army Pamphlets& 135  & 7395589  & 9\%  \\
FM      & Field Manuals                   & 55   & 6159531  & 7\%  \\
STP     & Soldier Training Publications   & 40   & 4581602  & 6\%  \\
TB      & Technical Bulletins             & 39   & 1642999  & 2\%  \\
ADP     & Army Doctrine Publications      & 16   & 823817   & 1\%  \\
AD      & Army Directives                 & 117  & 346724   & <1\% \\
MISC    & Types w/ < 10 Examples          & 16   & 276230   & <1\% \\
ALARACT & All Army Activities             & 69   & 180762   & <1\% \\
SD      & Strategic Documents             & 12   & 122196   & <1\% \\ \midrule
        & \multicolumn{1}{r}{\textbf{Total:}} & 4329 & 82492464 & 100\% \\ 
\bottomrule
\end{tabular}
\end{table}

\subsection{TRACLM-v3 Generated Q\&A Examples by Category}

For each question category, the prompt used to generate Q\&A training examples and a single high-quality generated example are provided. All questions were generated with the following system prompt that preceded the category-specific prompt:

\noindent \textbf{Guidelines:}
\begin{Verbatim}
### GUIDELINES
While performing the task, strictly adhere to the following guidelines:
1. Do NOT add new information or alter the factual details of the input; rely solely on the information in the provided text.
2. Do NOT respond with anything other than a single question and answer pair in the format above.
3. DO NOT MENTION THE "input" or "text" in your questions or answers (e.g. do not say "based on the text", "based on the provided document", "per the input", etc.)
4. Seek to maximize the amount of domain-specific information present in the question and answer pairs. Longer questions & answers are favorable if you go out of your way to include more domain-specific terms, acronyms, and concepts.
\end{Verbatim}

\subsubsection{Summarization}

\noindent \textbf{Prompt}:
\begin{Verbatim}
You are a helpful AI assistant acting as an expert-level tutor for a U.S. military science course. Help me create quiz questions that will help me study for an upcoming exam.

Given an unstructured block of text from a U.S. Army publication, generate a question and answer that evaluates the reader's ability to summarize the text to maximize comprehension.

Here is an example:

### INPUT EXAMPLE
[achieving strategic objectives through direct military confrontation with the United States. Therefore, adversaries pursue their objectives indirectly through malign activities and armed conflict, targeting others in ways calculated to avoid war with the United States. These activities include subversive political and legal strategies, establishing physical presence on the ground to buttress resource claims, coercive economic practices, supporting proxy forces, and spreading disinformation. However, several adversaries have both the ability and the will to conduct armed conflict with the United States under certain conditions. This requires Army forces to be prepared for limited contingencies and large-scale combat operations. 2-4. Global and regional adversaries apply all instruments of national power to challenge U.S. interests and the joint force. Militarily, they ha]

### OUTPUT EXAMPLE
Q: Provide a summary of the input text regarding how adversaries pursue their objectives.
A: Adversaries pursue their objectives indirectly through malign activities and armed conflict, targeting others in ways calculated to avoid war with the United States. These activities include subversive political and legal strategies, establishing physical presence on the ground to buttress resource claims, coercive economic practices, supporting proxy forces, and spreading disinformation. Several adversaries have both the ability and the will to conduct armed conflict with the United States under certain conditions, requiring Army forces to be prepared for limited contingencies and large-scale combat operations. Global and regional adversaries apply all instruments of national power to challenge U.S. interests and the joint force.
\end{Verbatim}

\newpage

\noindent \textbf{Example}:
\begin{Verbatim}
['What is the role and management structure of the Army Fire and Emergency Services (FES) program, as well as its key programs and procedures for deployed locations and contract safety requirements?'
 'The Army Fire and Emergency Services (FES) program provides structural, aircraft fire and rescue, wildland fire response, and control and extinction of fires when necessary to protect people, property, equipment, or mission capability. It also provides first response capabilities to fire-related hazmat/weapons of mass destruction incidents in an all-hazards environment. The FES program is managed at the higher headquarters programmatic level by the DCS, G 9, who is also the Authority Having Jurisdiction (AHJ) for the U.S. Army. At the installation/activity level, management and administration of FES should involve effective communication and collaboration between the garrison commander and the installation fire chief.\n\nThe key FES programs include management, fire prevention services, fire safety deficiencies, risk assessment coding, FES training, aircraft rescue firefighting, EMS, bloodborne pathogens, emergency dispatch services, wildland firefighting, hazmat, fire protection systems, water distribution systems, weapons of mass destruction, and CBRNE.\n\nFor deployed locations, the installation fire chief will meet with the local safety office on a regular basis to collaborate and coordinate on the FES programs and services as applicable for the local installation. Firefighting involving munitions at deployed locations will adhere to specific procedures, such as advanced planning, evacuation plans, and required withdrawal distances.\n\nContract safety requirements state that contracts will require that contracted activities be conducted in a manner that minimizes the potential for mishaps, impacts on army operations or critical assets, and unacceptable risk to human health and the environment. Contractors must comply with contract SOH requirements and applicable federal, state, and local laws, regulations, codes, and standards, including SOH requirements.']
\end{Verbatim}

\newpage

\subsubsection{Domain Vocabulary}

\noindent \textbf{Prompt}:
\begin{Verbatim}
You are a helpful AI assistant acting as an expert-level tutor for a U.S. military science course. Help me create quiz questions that will help me study for an upcoming exam.
Given an unstructured block of text from a U.S. Army publication, generate a question and answer that measures the reader's ability to correctly use domain-specific vocabulary. 

Here is an example:

### INPUT EXAMPLE
[tion strategy and program. e. TSS facilities resourcing are influenced by the Army Infrastructure Enterprise as part of the Army’s overall facilities strategy. 5–2. Planning, programming, budgeting and execution a. PPBE is the Army’s primary resource management and decisionmaking process, the PPBE interfaces with OSD and Joint planning and links directly to OSD programming and budgeting. It develops and maintains the Army portion of the Defense program and budget. It supports Army planning, and it supports program development and budget preparation at all levels of command. It supports execution of the approved program and budget by both HQ and field organizations. During execution, it provides feedback to the PPBE processes. b. The PPBE ties strategy, program, and budget together. It helps build a comprehensive plan in which budgets flow from programs, programs from requirements, requirements from missions, and missions from national security objectives. The patterned flow from end purpose to resource cost defines requirements in progressively greater details. (1) Long range planning creates a vision of the Army 10 to 20 years into the future. (2) In the 2- to 15-year midterm, long range macro estimates give way to a specified size, composition, and quality of divisional and support forces. (3) In the 0- to 2-year near term, budgeting converts program requirements into requests for manpower and dollars. c. When enacted into appropriations and manpower authorizations, these resources become available. d. TSS resourcing occurs in accordance with AR 1–1. The PPBE and other management events impact TSS resourcing. Although minor slippage of these events occurs with some frequency, DCS, G–3/5/7 (G–37/TR) will use the PPBE management timeline for planning and to time decisions and associated outputs that may impact available resources. e. HQDA is continually involved in financial planning and execution for the current FY, budgeting for the next FY, and programming for five POM years. Because financial planning is a constant, rolling cycle, the POM “lock” associated with each FY translates to the budget estimate submission. When the first year of each POM is translated to the budget estimate submission, new requirements for that FY are no longer accepted. Instead, the new requirement or “add” must be worked as part of the next POM build. For requirements to receive funding, submittals must take place per with the POM and management timelines. f. Core resourcing is contained in key training and installation management programs. 5–3. Management Decision Packages a. Early in the PPBE process, the resource management architecture distributes program and budget resources to MDEPs by appropriation and program element. MDEPs serve as a resource mana]
\end{Verbatim}

\newpage

\begin{Verbatim}
### OUTPUT EXAMPLE
((Q)): What is the primary resource management and decisionmaking process for the Army?
((A)): The primary resource management and decisionmaking process for the Army is the Planning, Programming, Budgeting, and Execution (PPBE) process. The PPBE interfaces with OSD and Joint planning and links directly to OSD programming and budgeting. It develops and maintains the Army portion of the Defense program and budget, supports Army planning, and supports program development and budget preparation at all levels of command. It supports execution of the approved program and budget by both HQ and field organizations, providing feedback to the PPBE processes during execution. The PPBE ties strategy, program, and budget together, helping build a comprehensive plan in which budgets flow from programs, programs from requirements, requirements from missions, and missions from national security objectives. The patterned flow from end purpose to resource cost defines requirements in progressively greater details. Long range planning creates a vision of the Army 10 to 20 years into the future, while in the 2- to 15-year midterm, long range macro estimates give way to a specified size, composition, and quality of divisional and support forces. In the 0- to 2-year near term, budgeting converts program requirements into requests for manpower and dollars. When enacted into appropriations and manpower authorizations, these resources become available. TSS resourcing occurs in accordance with AR 1–1, with the PPBE and other management events impacting TSS resourcing. HQDA is continually involved in financial planning and execution for the current FY, budgeting for the next FY, and programming for five POM years. Core resourcing is contained in key training and installation management programs.
\end{Verbatim}

\noindent \textbf{Example}:
\begin{Verbatim}
['Under what circumstances does a student-applicant not meet the dependency requirements for the SROTC program, and what measures must be taken if an applicant or cadet becomes a sole parent?'
 "A student-applicant does not meet the dependency requirements for the SROTC program if they are married with three or more dependents under 18 years of age, if they are married and their spouse is in the SROTC program or any of the U.S. armed forces (excluding the Individual Ready Reserve) and have one or more dependents under 18 years of age, or if they are an unmarried, divorced, or sole parent who has custody of one or more dependents under 18 years of age. If an applicant is not married and their child or children have been placed in the custody or guardianship of another by court order or a valid written custody agreement with no child support requirement, they are eligible to enroll without a waiver. If an applicant or cadet becomes a sole parent for any reason, they must execute a family care plan, and disenrollment may result if no such plan is implemented. Additionally, if the number, status, or circumstances of a cadet's dependents adversely affect their performance, they may be processed for disenrollment. Green-to-Gold Active Duty Option Program (G2GADOP) applicants will require a dependency waiver, including sole and joint custody parents. All applicants whose primary language is other than English will be screened, and those in Puerto Rico who score below 90 must enroll in the English as a Learned Language program, while applicants in all other locations must improve their score to 90 or better by enrolling in a language program at their college, the Defense Language Institute, or in their local area."]
\end{Verbatim}

\subsubsection{Natural Language Inference}

\noindent \textbf{Prompt}:
\begin{Verbatim}
You are a helpful AI assistant acting as an expert-level tutor for a U.S. military science course. Help me create quiz questions that will help me study for an upcoming exam.
Given an unstructured block of text from a U.S. Army publication, generate a question and answer that assesses the reader's domain-specific knowledge through natural language inference. 

Here is an example:

### INPUT EXAMPLE
[Chapter 2 Framework  2–1. Army Leader Development Program framework This chapter outlines the framework used to execute the ALDP. 2–2. Army Leader Development Process Multiple stakeholders are involved in shaping strategic decisions that impact Army leader development. The Army Leader Development Process Model communicates fundamental leader development authorities, roles, stakeholder organizations, and forums. Proposed policy changes flow top to bottom and bottom to top, while synchronization and integration occur between each of the organizations and forums. Figures 2–1 and 2–2 are simple depictions of the boundaries and interface between key stakeholder organizations. It is essential to understand— a. The Secretary of the Army has statutory authority to administer Title 10 (10 USC) functions on behalf of the Army and vests authority in the Assistant Secretary of the Army (Manpower and Reserve Affairs) (ASA (M&RA)) to execute their responsibilities. b. The Chief of Staff, Army (CSA) leads and shapes leader development for the Army. c. The ASA (M&RA) is responsible for supervising the development of all policies, plans, and programs pertaining to readiness resourcing, training, and professional and leader education and development. d. The Deputy Chief of Staff, G–3/5/7 (DCS, G–3/5/7) is the Army staff (ARSTAF) proponent for Army training and leader development policy and resourcing. e. The Commanding General, U.S. Army Training and Doctrine Command (CG, TRADOC) is the senior responsible official (SRO) for Army leader development. f. The Army Leader Development Process Model depicts the various roles and responsibilities for leader development by forum (see fig 2–1). The matrix represents the four levels of authority within the Army Leader Development Process: 2 DA PAM 350–58 • 8 March 2013 Figure 2–1. Army Leader Development Process Model (1) The ASA (M&RA)]

### OUTPUT EXAMPLE
((Q)): Given the premise that the Secretary of the Army has the statutory authority to administer Title 10 functions on behalf of the Army, and the hypothesis that the Chief of Staff of the Army leads and shapes leader development for the Army, does the premise entail the hypothesis? 
((A)): No, just because the Secretary of the Army has Title 10 statutory authority, this does not guarantee the Chief of Staff of the Army will lead and shape leader development for the Army. The Secretary of the Army has statutory authority to administer Title 10 functions on behalf of the Army and vests authority in the Assistant Secretary of the Army (Manpower and Reserve Affairs) to execute their responsibilities. Regardless, the Chief of Staff of the Army, who works directly for the secretary, does indeed lead and shape leader development for the Army.   
\end{Verbatim}

\noindent \textbf{Example}:
\begin{Verbatim}
['Given the premise that a Marine Expeditionary Brigade (MEB) is not structured to conduct offensive tasks as a brigade but it may plan, conduct, and provide command and control for offensive tasks performed by assigned maneuver units and a Tactical Command Force (TCF) as part of security for the support area, and the hypothesis that the MEB can form task forces or company teams to support the offensive operations of its supported headquarters, does the premise entail the hypothesis?'
 'Yes, the premise does entail the hypothesis. The MEB can plan, conduct, and provide command and control for offensive tasks performed by assigned maneuver units and a TCF as part of security for the support area, and it can also form task forces or company teams to support the offensive operations of its supported headquarters. These task forces or company teams can be attached or placed operational control (OPCON) to battalion combat teams (BCTs) or employed by the MEB to complement or reinforce maneuver forces across the area of operation (AO) of higher headquarters for a specific or select mission or tasks that support the main effort. However, specific or select missions do not imply long-term task organization. If the MEB becomes a force provider for an enduring period of time, it could become ineffective in its ability to conduct support area operations.']
\end{Verbatim}

\subsubsection{Commonsense Reasoning}

\noindent \textbf{Prompt}:
\begin{Verbatim}
You are a helpful AI assistant acting as an expert-level tutor for a U.S. military science course. Help me create quiz questions that will help me study for an upcoming exam.
Given an unstructured block of text from a U.S. Army publication, generate a question and answer that tests the reader's ability to perform commonsense reasoning with the domain-specific information. 

Here is an example:

### INPUT EXAMPLE
[nto a security force, a main body, and a reserve, all supported by sustainment organizations. The best place and time for an attacking force to task-organize is when it is in an assembly area. This allows units to complete  any changes in task organization in time to conduct rehearsals with their attached and supporting elements. FORWARD SECURITY FORCE 5-3. While planning and preparing for operations, units have a security force to their front. Upon initiating movement toward their objective, they place a reconnaissance and security force to their front to identify enemy locations, dispositions, and strengths. These forces also confirm trafficability of axes of advance and cross-mobility corridors. They also destroy (within capabilities) as much of the enemy in the disruption zone as possible, enabling the main body to focus on the enemy in the battle zone. Units only resource dedicated flank or rear security forces during an attack if the attack uncovers one or more flank or the rear of the attacking force as they advance. Commanders designate flank or rear security forces and assign them a guard or screen mission, depending on the mission variables. Attacking forces should maintain a forward security element. The size of the forward security element is based upon if the friendly force has gained and maintained visual contact of the enemy. MAIN BODY 5-4. Units organize their main bodies into combined arms formations to conduct their main and supporting efforts. They aim their main effort towards]

### OUTPUT EXAMPLE
((Q)): To what extent should the forward security force destroy the enemy in the disruption zone and why?
((A)): The forward security force should destroy as much of the enemy in the disruption zone as possible, within their capabilities. This enables the main body to focus on the enemy in the battle zone.    
\end{Verbatim}

\noindent \textbf{Example}:
\begin{Verbatim}
['In the context of the Rear Detachment (RD) concept, what activities can Chaplain Sections and Unit Ministry Teams (UMTS) execute during the "Establish Security and Restore Services" phase to support the force and prepare for redeployment?'
 'During the "Establish Security and Restore Services" phase, Chaplain Sections and UMTS can provide and advise field or consolidated worship based on the mission variables, offer counseling and casualty care, conduct memorials, provide redeployment classes, continue integration into the staff operations processes, and focus on advisement and consolidation of Religious Support (RS) provision. They also prepare for re-supply, relief in place or transfer of authority, handle rs activities such as area coverage responsibilities, task organization changes, facilities, and religious factors information, and prepare for redeployment.']    
\end{Verbatim}

\subsubsection{Paraphrase Detection}

\noindent \textbf{Prompt}:
\begin{Verbatim}
You are a helpful AI assistant acting as an expert-level tutor for a U.S. military science course. Help me create quiz questions that will help me study for an upcoming exam.
Given an unstructured block of text from a U.S. Army publication, generate a question and answer that asks readers to compose new text that meaningfully supports or contradicts the provided input to improve their in-domain reasoning ability. 

Here is an example:

### INPUT EXAMPLE
[ndirect fires, or improvised explosive devices and may be against stationary or moving forces. The design of the landscape, coupled with climatic conditions, creates a unique set of mountain operations characteristics that are characterized by close fights with dismounted infantry, decentralized small-unit operations, degraded mobility, increased movement times, restricted lines of communications, and operations in thinly populated areas. (Refer to ATTP 3-21.50 for additional information.) Desert Terrain  1-15. Operations in desert terrain require adaptation to the terrain and climate. Equipment must be adapted to a dusty and rugged landscape with extremes in temperature and changes in visibility. The battalion orients on primary enemy approaches but prepares for an attack from any direction. Considerations for operations in desert terrain include lack of concealment and the criticality of mobility; use of obstacles to site a defense, which are limited; strong points to defend choke points and other key terrain; and mobility and sustainment. (Refer to FM 90-3 for additional information.) Jungle Terrain  1-16. Operations in jungle terrain combine dispersion and concentration. For example, a force may move out in a dispersed formation]

### OUTPUT EXAMPLE
((Q)): Compose a new paragraph that contradicts the characteristics of mountain operations.
((A)): Mountain operations are not primarily characterized by close engagements with dismounted infantry or decentralized small-unit operations. Instead, they often involve long-range engagements facilitated by the elevated vantage points that mountains provide. Mobility is not significantly degraded; rather, the unique topography of mountainous areas can offer alternative routes and paths for maneuvering, enhancing mobility options. Movement times are not necessarily increased; in some cases, the use of specialized mountain equipment and training can lead to efficient navigation and rapid troop movements.
   
\end{Verbatim}

\noindent \textbf{Example}:
\begin{Verbatim}
['Compose a new paragraph explaining how the process of approval for actual expense allowance has changed, specifically emphasizing the reduction in the number of approval authorities required.'
 'Previously, under the guidelines set forth in reference 1.c., requests for actual expense allowance needed to be approved by an authority one organizational level above the requester. However, due to the recent changes detailed in references 1.a., 1.b., and 1.d., this requirement has been modified. Now, appointed order-issuing officials hold the approval authority for actual expense allowance, allowing for more streamlined decision-making when it comes to reimbursing travelers for expenses that surpass the maximum locality rate, up to 300% of the locality per diem rate, in exceptional situations. Consequently, the role of the Administrative Assistant to the Secretary of the Army as an approval authority for such requests has been eliminated, unless specifically appointed for this task. This adjustment aims to enhance efficiency in the approval process for actual expense allowances related to travel by Department of the Army senior officials.']   
\end{Verbatim}

\subsubsection{Completion}

The completion category differed from the above in that it did not utilize synthetic examples generated from a larger LLM. Instead, completion examples were created by randomly selecting from a list of simple questions (e.g. "How would you complete this text?", "How to continue this?", etc), dividing the given chunk at the 25\%, 50\%, or 75\% mark, and inserting elements into the prompt as appropriate. See below for a full example:

\noindent \textbf{Example}:
\begin{Verbatim}
['What is the most fitting way to complete this text?
### INPUT
chapter 1 general 11. purpose this regulation contains policies, and procedures for using the polygraph in the department of the army (da). it prescribes the authority for conducting polygraph examinations within da. it delineates responsibility and authority between u.s. army criminal investigation command (usacidc) and u.s. army intelligence and security com- mand (inscom). policy regarding the intelligence polygraph program, including counterintelligence scope poly- graph examinations, is covered in ar 381  20(c). 12. references and forms see appendix a. 13. explanation of abbreviations and terms see the glossary. 14. responsibilities a. the deputy chief of staff, g  2 will (1) designate the director, army g  2x to centralize technical control...
### OUTPUT
...nal action of an investigation. a. questions asked will relate to the relevant issue under inquiry. ar 1956  21 april 2016 2 b. unless directly relevant to the inquiry, the following issues will not be covered (1) religious beliefs and lawful affiliations. (2) beliefs and opinions regarding racial matters. (3) political beliefs and affiliations of a non-subversive nature. (4) opinions on the constitutionality of legislative policies. (5) sexual orientation or preferences of subject. c. for criminal investigations, no adverse action will be taken against persons refusing or unwilling to be examined by polygraph. other investigation, as appropriate, may continue following the person's refusal to be examined by polygraph. d. army intelligence polygraph examinations, including counterintelligence scope polygraph (csp) examinations of persons with access to certain highly sensitive and classified programs, are governed by the provisions in ar 381  20(c). e. dod requests for polygraph examinations will be submitted to the commander, usacidc. for us]
\end{Verbatim}

\newpage

\subsection{Additional TRACLM Output Comparisons}

\begin{figure}[H]
    \centering
    \includegraphics[width=1\textwidth]{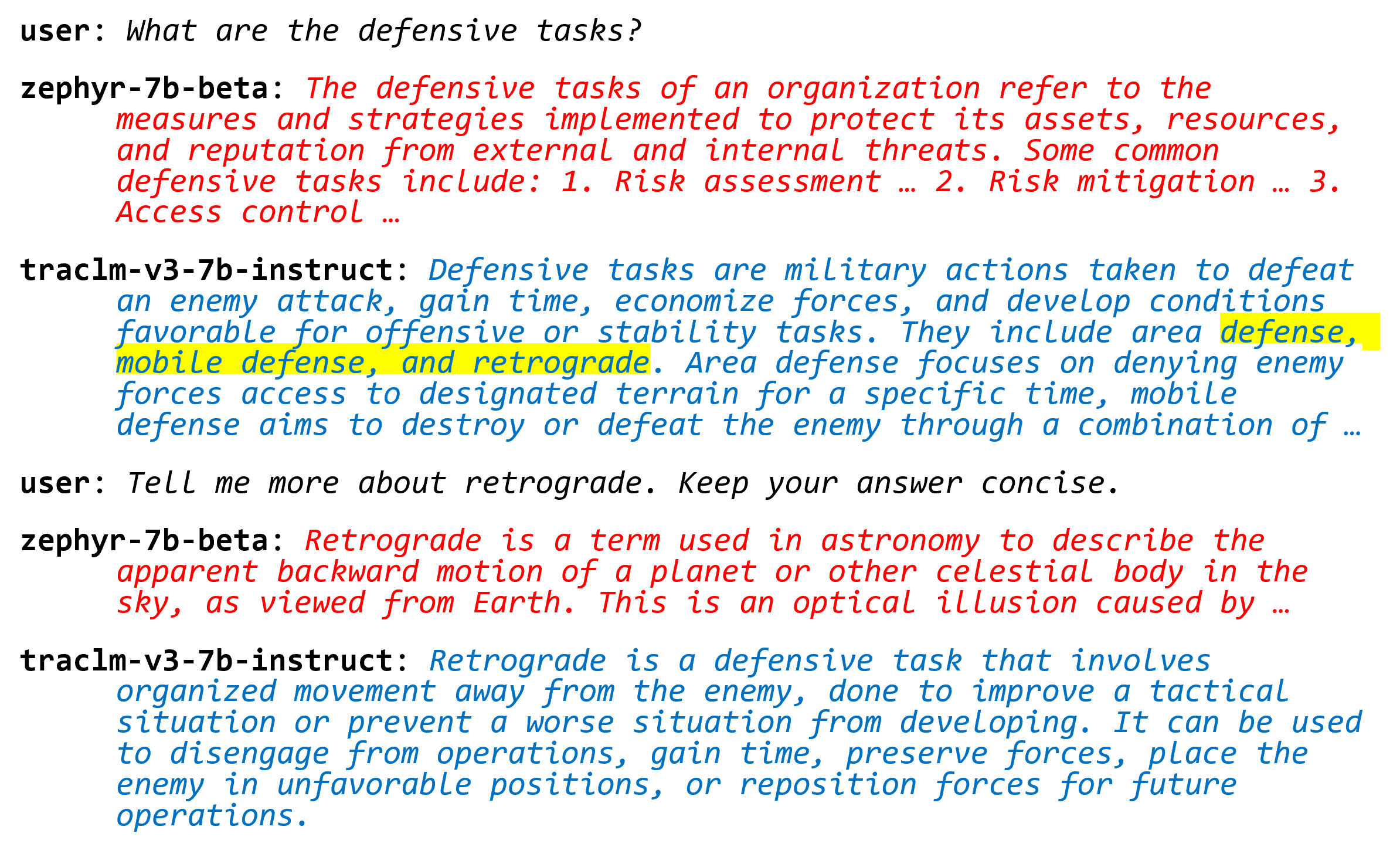}
    \caption{Zephyr-7b-beta vs. TRACLM-v3 Response Quality, Exhibit B}
    \label{fig:defensive_tasks}
\end{figure}

\begin{figure}[H]
    \centering
    \includegraphics[width=1\textwidth]{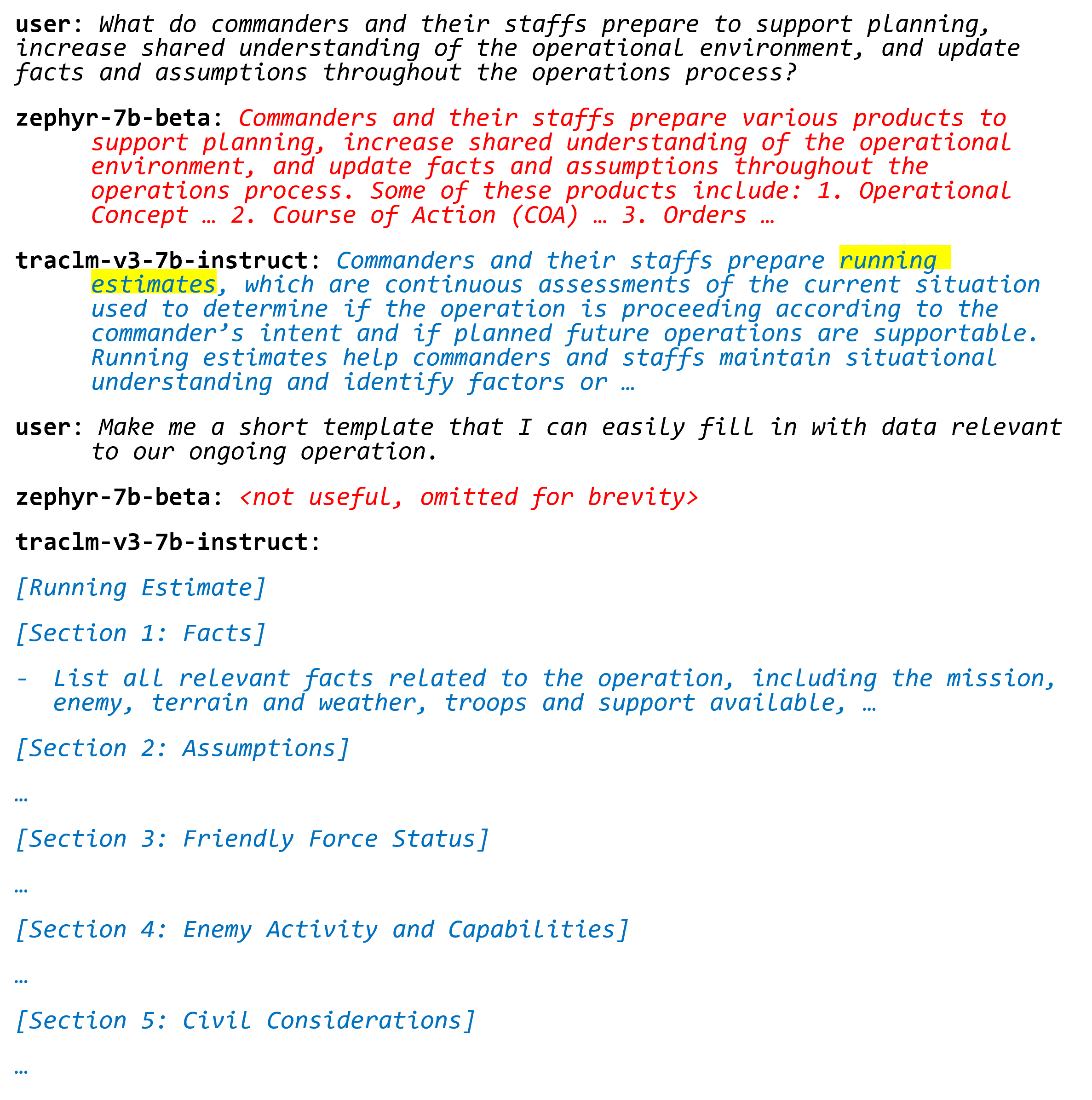}
    \caption{Zephyr-7b-beta vs. TRACLM-v3 Response Quality, Exhibit C}
    \label{fig:running_estimates}
\end{figure}

Figures \ref{fig:defensive_tasks} \& \ref{fig:running_estimates} are additional examples of simultaneous interactions with a general-purpose open-source LLM and TRACLM-v3. In both cases, TRACLM-v3 exhibits clearly improved domain-specific knowledge in its responses.

\newpage

\newpage

\subsection{MilGLUE Dataset Curation}

\subsubsection{Topic Modelling Seed Topic Keywords}

Table \ref{tab:milglue_topic_seeds} lists the keywords that were chosen to represent eight seed topics that were used by BERTopic to group the MILGLUE data in preparation for sampling. The keywords were chosen from a combination of keywords from topics that BERTopic discovered without seed topics and keywords manually chosen by  a human.

\label{tab:milglue_topic_seeds}
\begin{table}[H]
\centering
\caption{Seed topic keywords provided to BERTopic.}
\begin{tabular}{@{} l p{0.8\linewidth} @{}}
\toprule
\textbf{Topic} & \textbf{Seed Topic Keywords} \\ \midrule
Topic\_0 & leadership, administration, rules, regulations, judge, spends         \\ 
Topic\_1 & military operations, organization, fires, attack, ballistic, deterrent, jungle, fish, poisonous, barracuda \\ 
Topic\_2 & equipment, technology, helicopter, aircraft, airplane, boat, ship, vehicle, missile, cruise, warhead, stryker \\ 
Topic\_3 & training, education, simulation, doctrine                            \\
Topic\_4 & logistics, supply chain, BSB                                       \\ 
Topic\_5 & strategy, tactics, intelligence, reconnaissance, joint, jtf          \\
Topic\_6 & personnel, recruitment, retirement, health, safety, workforce       \\ 
Topic\_7 & international, foreign, affairs, relations, diplomacy                \\
\bottomrule
\end{tabular}
\end{table}

\subsubsection{Sentence Similarity (Binary)}

\label{sec:MilGLUE_Dataset_Makeup}
Below are tables showing the topics found by the BERTopic model, and how many records were identified within each topic. Topics are given a name based on common keywords that occur in the topic. Topic -1 represents outliers that could not be grouped with other topics.

\begin{table}[H]
\centering
\caption{Topic breakdown of the curated Sentence Similarity (Binary) MilGLUE dataset.}
\begin{tabular}{@{} l r @{}}
\toprule
\textbf{Topic} & \textbf{Records} \\ \midrule
-1\_organizations\_army\_organized\_understand & 308 \\
\textbf{0\_operations\_support\_airspace\_multinational} & \textbf{747} \\
1\_doctrine\_changes\_change\_knowledge & 746 \\
2\_module\_11\_rmy\_upport & 746 \\
3\_message\_format\_aircraft\_variable & 746 \\
4\_responsiveness\_fluid\_independence\_optimize & 746 \\
5\_line\_fuel\_quadripartite\_equally & 746 \\
6\_small\_surrounding\_feet\_msl & 701 \\
7\_distribution\_supply\_transit\_visibility & 746 \\
8\_state\_law\_sending\_receiving & 746 \\
9\_things\_technology\_ying\_tactic & 674 \\
10\_letter\_designating\_latitude\_earth & 349 \\
11\_messages\_corresponding\_list\_series & 356 \\
12\_hart\_memoirs\_london\_liddell & 317 \\
13\_laundry\_shower\_decontamination\_weekly & 314 \\
14\_nfusion\_keypad\_gars\_anatomical & 293 \\
15\_white\_night\_illumination\_goggle & 200 \\
16\_tais\_versions\_hmtl\_device & 178 \\
17\_lethal\_sent\_report\_interrogator & 172 \\
18\_manned\_corrected\_barometric\_amsl & 169 \\
\midrule
\textbf{Total} & \textbf{10,000} \\
\bottomrule
\end{tabular}
\label{tab:records_per_topic_ssb}
\end{table}

\subsubsection{Paraphrase}

\begin{table}[H]
\centering
\caption{Topic breakdown of the curated Paraphrase MilGLUE dataset.}
\begin{tabular}{@{} l r @{}}
\toprule
\textbf{Topic} & \textbf{Records} \\ \midrule
\textbf{-1\_operations\_support\_units\_information} & \textbf{4585} \\
\textbf{0\_operations\_support\_units\_staff} & \textbf{4585} \\
1\_contracting\_contract\_contracts\_contractors & 182 \\
2\_cpof\_tool\_shared\_used & 129 \\
3\_election\_bosnia\_osce\_sfor & 68 \\
4\_aircraft\_imagery\_avim\_maintenance & 65 \\
5\_security\_afghanistan\_afghan\_peacekeeping & 59 \\
6\_band\_music\_mpt\_musical & 54 \\
7\_interpreters\_language\_interpreter\_linguist & 49 \\
8\_drivers\_training\_driving\_vehicle & 43 \\
9\_water\_sanitation\_bottled\_fuel & 36 \\
10\_weather\_heat\_cold\_swo & 34 \\
11\_products\_website\_search\_handbooks & 33 \\
12\_mail\_postal\_apo\_isb & 14 \\
13\_legal\_frago\_fragos\_review & 12 \\
14\_dogs\_mwd\_detection\_working & 11 \\
15\_weapon\_barrel\_weapons\_holster & 11 \\
16\_engagements\_performance\_amo\_leader & 10 \\
17\_stinger\_teams\_frequency\_monitored & 10 \\
18\_navigation\_psn\_gps\_11 & 10 \\ 
\midrule
\textbf{Total} & \textbf{10,000} \\
\bottomrule
\end{tabular}
\label{tab:records_per_topic_paraphrase}
\end{table}

\subsubsection{Next Sentence Recognition}

\begin{table}[H]
\centering
\caption{Topic breakdown of the curated Next Sentence Recognition MilGLUE dataset.}
\begin{tabular}{@{} l r @{}}
\toprule
\textbf{Topic} & \textbf{Records} \\ \midrule
\textbf{-1\_operations\_support\_units\_unit} & \textbf{4326} \\
0\_operations\_support\_force\_training & 4325 \\
1\_coa\_discussion\_course\_staff & 336 \\
2\_refugees\_ramadan\_displaced\_refugee & 142 \\
3\_rotor\_rotation\_blade\_landing & 140 \\
4\_water\_ice\_aircraft\_wash & 126 \\
5\_ebola\_blood\_malaria\_infectious & 117 \\
6\_graphics\_powerpoint\_screen\_screens & 91 \\
7\_cards\_smart\_products\_storage & 88 \\
8\_article\_wars\_journal\_published & 82 \\
9\_website\_search\_user\_engine & 59 \\
10\_software\_drives\_viruses\_virus & 54 \\
11\_sipr\_access\_problem\_siprnet & 19 \\
12\_terrain\_analysis\_products\_needed & 17 \\
13\_skin\_bites\_body\_methods & 16 \\
14\_bandwidth\_throughput\_network\_900kbps & 14 \\
15\_energy\_electricity\_grown\_purely & 14 \\
16\_sketches\_cards\_range\_sector & 12 \\
17\_cashing\_checks\_otc\_check & 11 \\
18\_motion\_video\_platforms\_unified & 11 \\
\midrule
\textbf{Total} & \textbf{10,000} \\
\bottomrule
\end{tabular}
\label{tab:records_per_topic_nsr}
\end{table}

\subsubsection{Masked Reasoning}

\begin{table}[H]
\centering
\caption{Topic breakdown of the curated Masked Reasoning dataset.}
\begin{tabular}{@{} l r @{}}
\toprule
\textbf{Topic} & \textbf{Records} \\ \midrule
\textbf{-1\_operations\_support\_fm\_army} & \textbf{2406} \\
0\_operations\_support\_information\_personnel & 2405 \\
1\_water\_habitat\_05\_edible & 2405 \\
2\_weather\_desert\_terrain\_river & 1842 \\
3\_knife\_shadow\_inch\_stone & 246 \\
4\_recording\_notes\_handwritten\_source & 146 \\
5\_stol\_engine\_thrust\_aircraft & 92 \\
6\_degree\_simulation\_subdivision\_categories & 78 \\
7\_dipper\_star\_stars\_big & 70 \\
8\_attachments\_tabs\_document\_edition & 62 \\
9\_repeat\_questions\_topic\_question & 48 \\
10\_color\_green\_monochrome\_colors & 41 \\
11\_tai\_nai\_horn\_mpt & 33 \\
12\_farmer\_fight\_examples\_calluses & 26 \\
13\_men\_nouns\_masculine\_exclusively & 24 \\
14\_divided\_quadrants\_keypads\_cgrs & 24 \\
15\_hide\_site\_border\_yourplanning & 16 \\
16\_folders\_compiled\_compiles\_target & 12 \\
17\_display\_sensors\_airborne\_real & 12 \\
18\_unclass\_secret\_1gb\_thumb & 12 \\
\midrule
\textbf{Total} & \textbf{10,000} \\
\bottomrule
\end{tabular}
\label{tab:records_per_topic_mr}
\end{table}

\subsection{MilBench Evaluation Harness Task Configuration Example}
\label{sec:MEH_Configuration}
\begin{lstlisting}[frame=single, caption={MMLU MilBench Task Configuration}, label={lst:mmlu-yaml}, language=yaml]
task: mmlu
dataset_path: cais/mmlu
dataset_subset: all
test_split: test
fewshot_split: dev
fewshot_config:
  sampler: first_n
  filter_column: subject
  num_fewshot: 5
doc_to_text: "{{question.strip()}}\nA. {{choices[0]}}\nB. {{choices[1]}}\nC. {{choices[2]}}\nD. {{choices[3]}}\nAnswer:"
doc_to_choice: ["A", "B", "C", "D"]
doc_to_target: answer
metadata:
  - version: "0.0.1"
\end{lstlisting}

\subsection{MilBench Server User Interface}

\subsubsection{Evaluation Audit Interface}

\begin{figure}[H]
    \centering
    \includegraphics[width=1\linewidth]{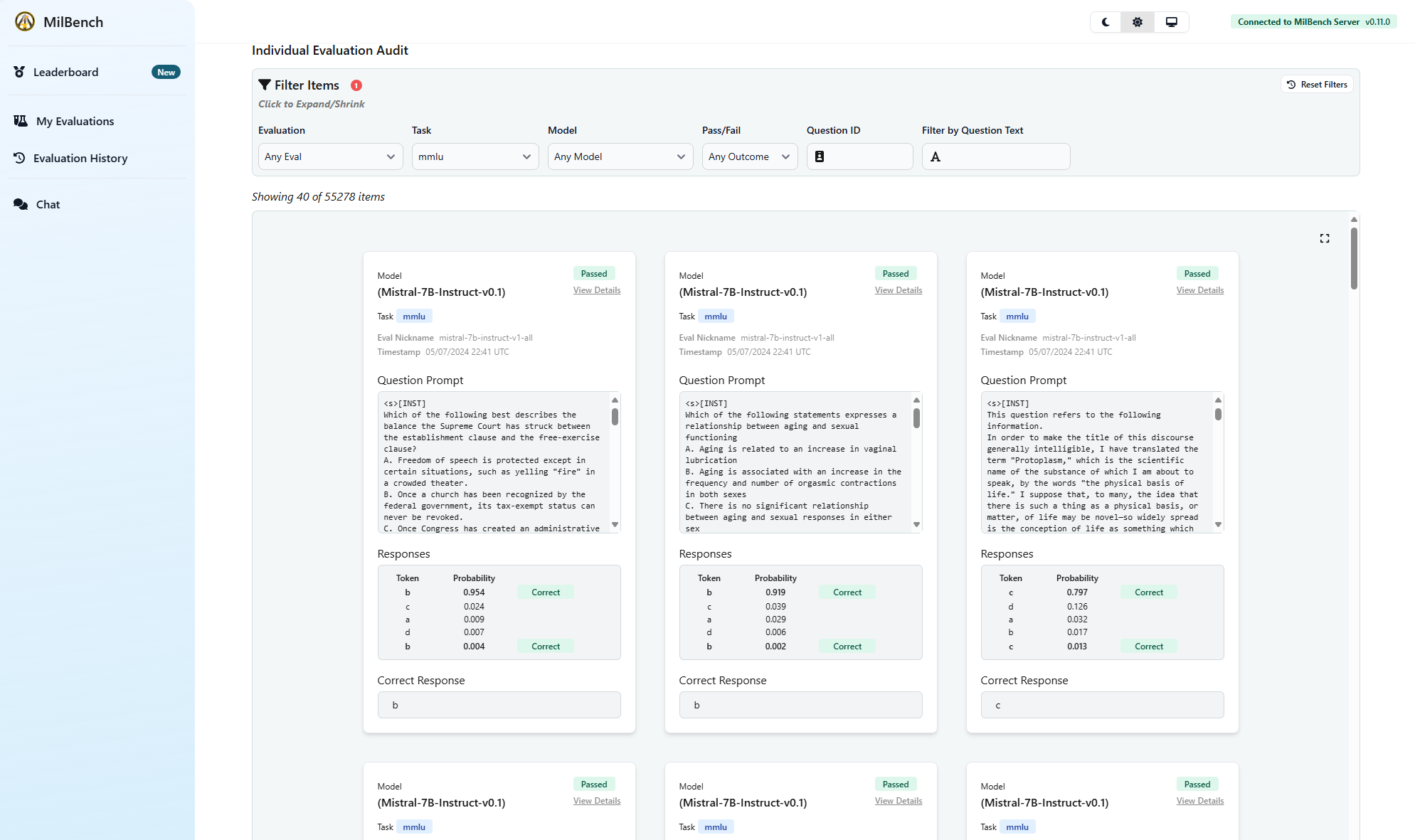}
    \caption{MilBench Server UI Evaluation Audit Interface}
    \label{fig:milbench_audit}
\end{figure}

Figure \ref{fig:milbench_audit} shows MilBench's evaluation audit interface. The interface provides question-level observably of evaluation results, even showing the probabilities of individual tokens within the LLM's response distribution. Results can be filtered by multiple criteria, including whether the LLM passed or failed a question.

\newpage

\subsubsection{MilBench LLM Leaderboard Interface}
\begin{figure}[H]
    \centering
    \includegraphics[width=1\linewidth]{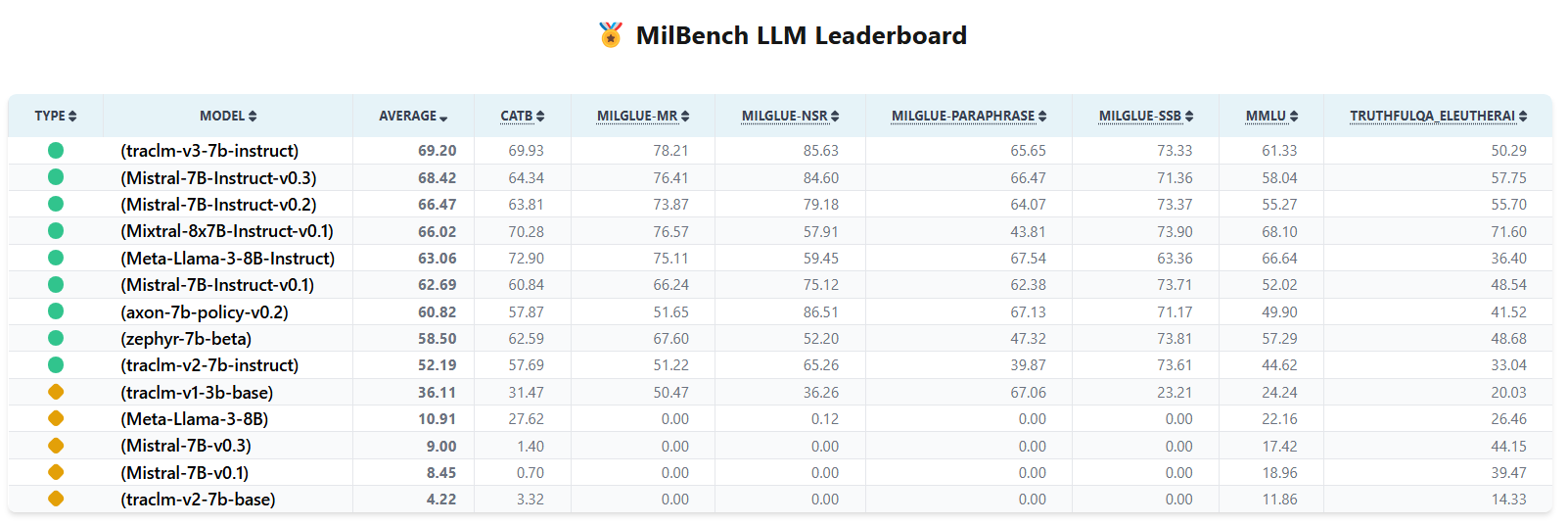}
    \caption{MilBench LLM Leaderboard}
    \label{fig:milbench_leaderboard}
\end{figure}

\begin{figure}[H]
    \centering
    \includegraphics[width=1\linewidth]{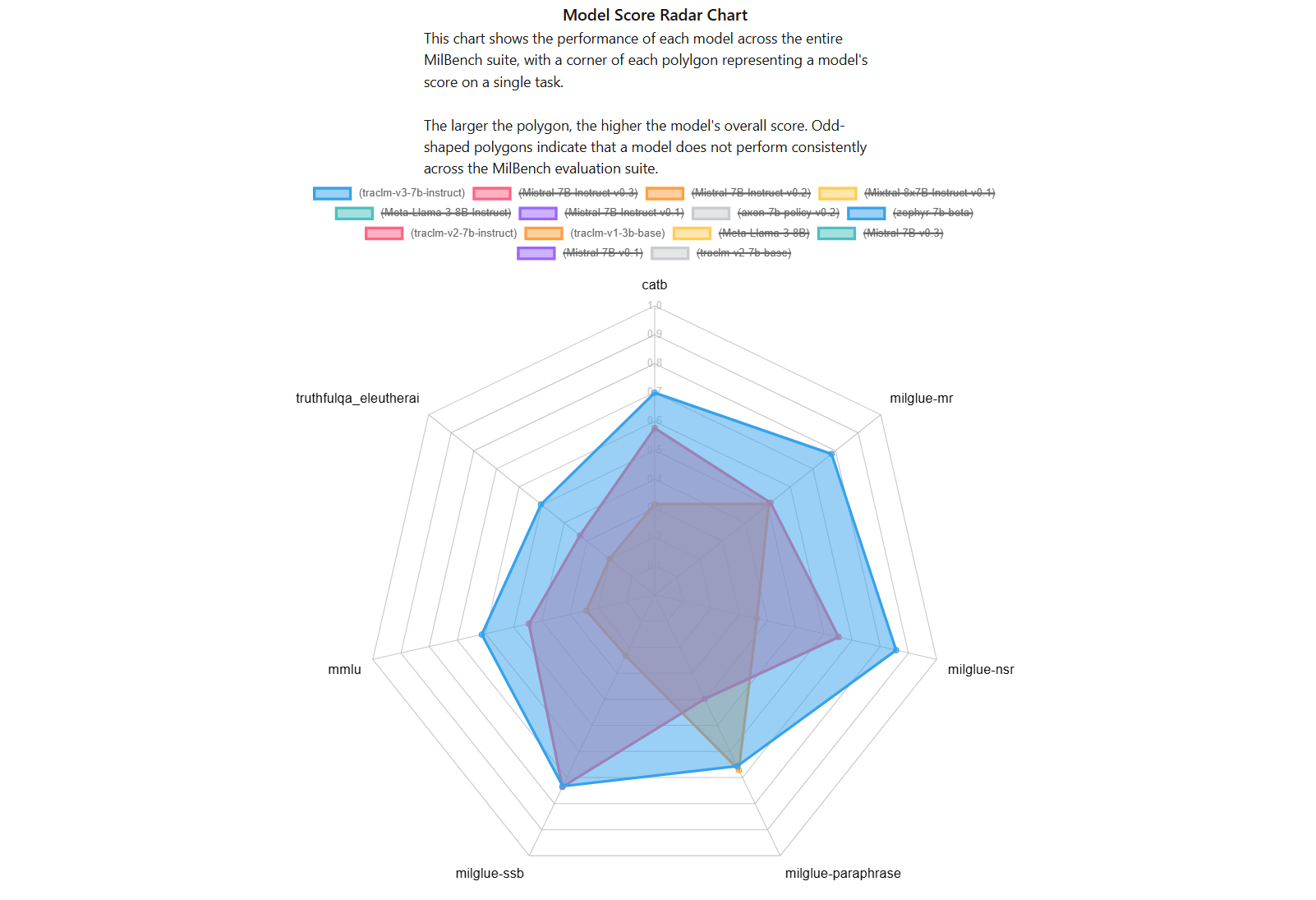}
    \caption{Interactive radar chart displays model performance across all evaluation tasks.}
    \label{fig:milbench_leaderboard_radar}
\end{figure}

Figures \ref{fig:milbench_leaderboard} and \ref{fig:milbench_leaderboard_radar} show the MilBench leaderboard and radar chart, respectively. The leaderboard is used to view detailed performance numbers for submitted models. Base (not fine-tuned) models are denoted by a orange diamond, while fine-tuned models are denoted by a green circle. The leaderboard can be interactively sorted, and hovering over a task header will show additional information about that task.

The radar chart provides a quick and intuitive understanding of a LLM's overall strengths and weaknesses. The colored layers representing each LLM's scores can be individually toggled by selecting the respective entry in the legend.

\subsubsection{MilBench Chat Evaluation Interface}
\label{fig:milbench_chat_evaluation}
\begin{figure}[H]
    \centering
    \includegraphics[width=1\linewidth]{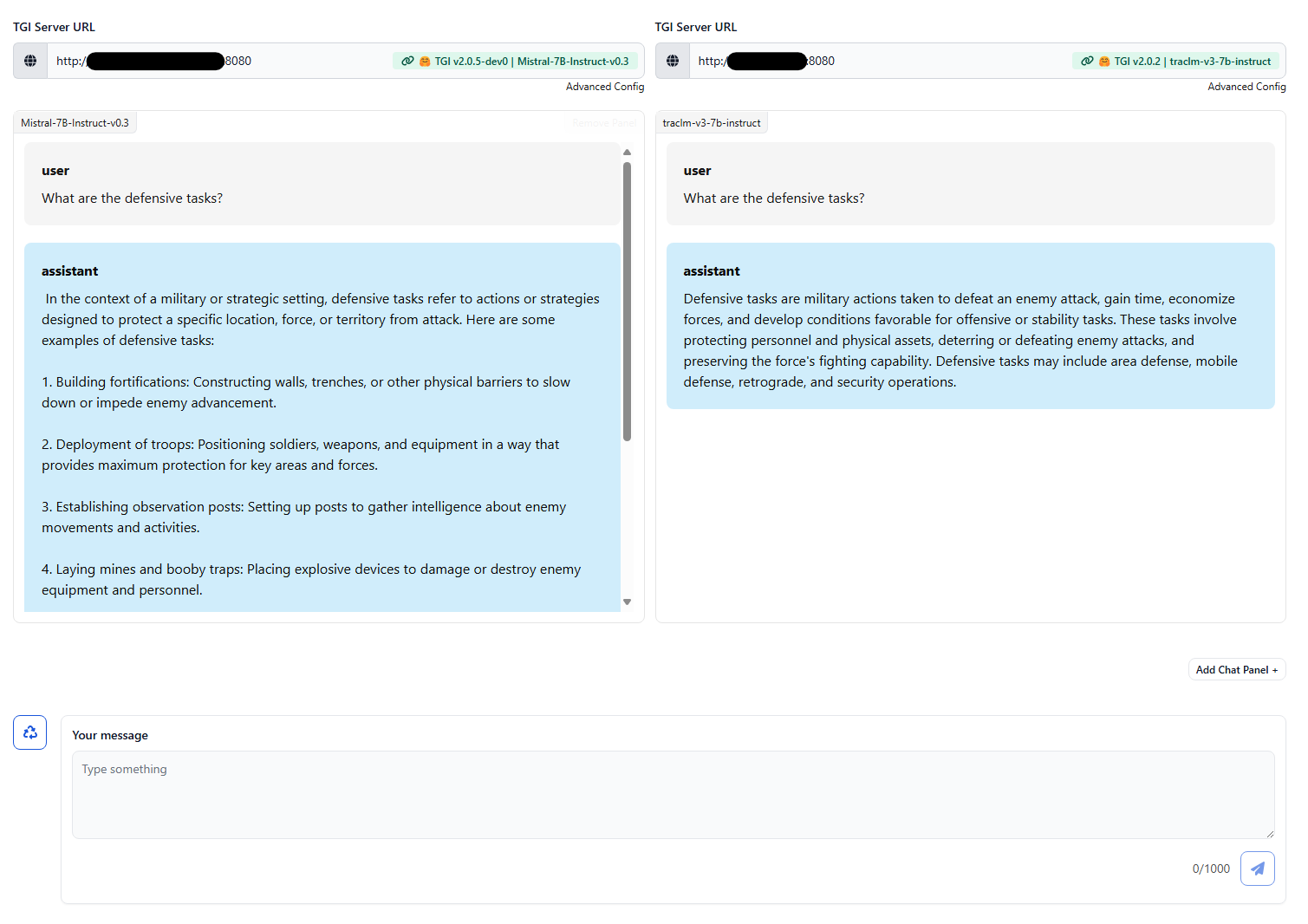}
    \caption{MilBench chat evaluation interface showing two models side-by-side responding to the same prompt.}
\end{figure}